\documentclass{article} 
\usepackage{iclr2027_conference,times}

\usepackage{amsmath,amsfonts,bm}

\def\eqref#1{equation~\ref{#1}}

\def\1{\bm{1}}

\DeclareMathAlphabet{\mathsfit}{\encodingdefault}{\sfdefault}{m}{sl}
\SetMathAlphabet{\mathsfit}{bold}{\encodingdefault}{\sfdefault}{bx}{n}

\usepackage{url}
\usepackage{wrapfig}
 \usepackage{graphicx}
\usepackage{booktabs}
\usepackage{multirow}
\usepackage{tabularx}
\usepackage{array}
\usepackage{makecell}
\usepackage{graphicx}
\usepackage{xcolor}
\usepackage{placeins}
\usepackage{algorithm}
\usepackage{algpseudocode}
\usepackage{amsmath,amssymb} \usepackage{bbm}
\usepackage{amsthm}
\usepackage{longtable}
\usepackage{xurl}
\providecommand{\afkeqref}[1]{(\ref{#1})}
\providecommand{\afkcode}[1]{{\footnotesize\nolinkurl{#1}}}

\usepackage{hyperref}
\usepackage{capt-of}
\usepackage{caption}

\usepackage{fvextra}
\usepackage[most]{tcolorbox}
\tcbuselibrary{breakable,skins}

\definecolor{skillborder}{RGB}{120,135,150}
\definecolor{skilltitle}{RGB}{238,242,246}

\newtcolorbox{skillbox}[1]{
  enhanced,
  breakable,
  title={#1},
  colback=white,
  colframe=skillborder,
  colbacktitle=skilltitle,
  coltitle=black,
  fonttitle=\bfseries,
  boxrule=0.45pt,
  arc=1.2mm,
  outer arc=1.2mm,
  left=7pt,
  right=7pt,
  top=6pt,
  bottom=6pt,
  toptitle=4pt,
  bottomtitle=4pt,
  before skip=7pt,
  after skip=9pt
}

\title{RoboHarn-Evo: Evolving Hierarchical Physical Knowledge for Self-Improving Robotic Manipulation}

\author{%
\textbf{Shifeng Bao\textsuperscript{1,3*}, Fanding Huang\textsuperscript{2*},
Yihan Lin\textsuperscript{1,3}, Youhe Feng\textsuperscript{1,3},} \\
\textbf{Guanlin Li\textsuperscript{1,3}, Chen Zhao\textsuperscript{1,3},
Yang Li\textsuperscript{1,3}, Jiawei He\textsuperscript{5}, Cheng Chi\textsuperscript{1,4$\ddagger$}, Jing Zhang\textsuperscript{1,4$\dagger$}} \\
\normalfont\textsuperscript{1}School of Information, Renmin University of China \\
\normalfont\textsuperscript{2}Tsinghua University \\
\normalfont\textsuperscript{3}Key Laboratory of Data Engineering and Knowledge Engineering, Beijing, China \\
\normalfont\textsuperscript{4}Engineering Research Center of Database and Business Intelligence, Beijing, China \\
\normalfont\textsuperscript{5}XYZ Embodied AI, Beijing, China \\[6pt]
{\hypersetup{hidelinks}\href{https://github.com/RUCKBReasoning/RoboHarn-Evo}{\textcolor[HTML]{348BD4}{\textbf{GitHub}}}}
}

\makeatletter
\g@addto@macro\@thanks{%
  \footnotetext[1]{Equal contribution.}%
  \footnotetext[2]{Corresponding author.}%
  \footnotetext[3]{Project Leader.}%
}
\makeatother

\iclrfinalcopy
\begin{document}

\maketitle
\lhead{}
\begin{figure}[H]
    \centering
    \includegraphics[width=0.9\textwidth]{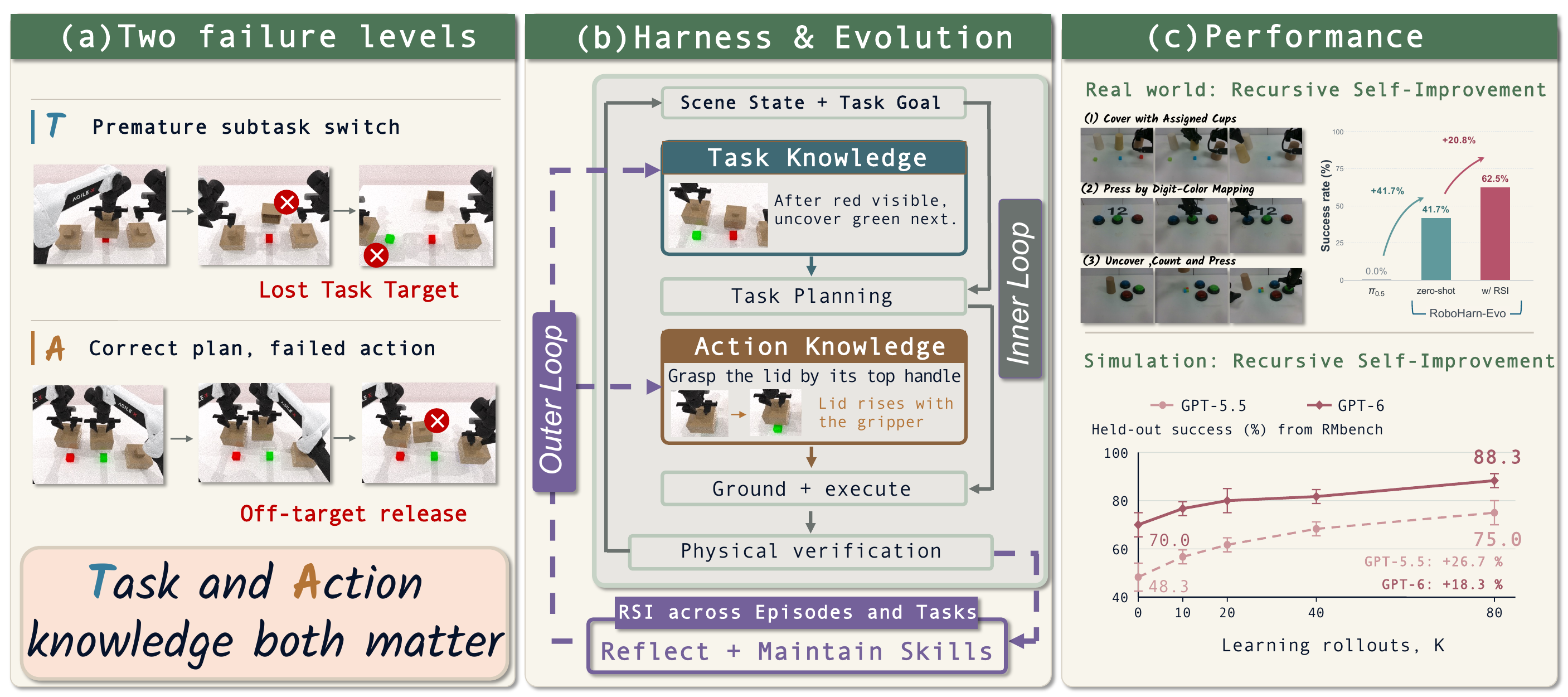}
    \caption{\textbf{Self-improving robotic manipulation through hierarchical knowledge evolution.}
  \textbf{(a)} In ordered block manipulation, the robot may switch to the next block before completing the current subtask (top), or select the correct
  target but release the cover off target (bottom).
  \textbf{(b)} Physical feedback updates Task and Action Knowledge across episodes, with the VLM and executor held fixed.
  \textbf{(c)} Knowledge evolution improves held-out simulation success by 18.3--26.7 percentage points after 80 rollouts. On real-world tasks,
  simulation-derived knowledge achieves 41.7\% mean zero-shot success, which rises to 62.5\% after further knowledge updates from real-world
  interaction.}
    \label{fig:teaser}
\end{figure}

\begin{abstract}
Vision-language models can coordinate long-horizon robot manipulation, yet successful task reasoning still depends on whether local physical interactions produce the intended effects. We study how repeated interaction can improve this capability without updating the base model. We introduce \textbf{RoboHarn-Evo}, a dual-loop harness that evolves \textbf{Hierarchical Physical Knowledge (HPK)} from physical experience. HPK couples two levels of reusable knowledge: Task Knowledge captures which subtask should be executed and when it is complete, while Action Knowledge
captures object-relative geometric strategies and their physical effects.
During execution, the agent retrieves knowledge at the corresponding decision
level and grounds it in the current scene under the task goal. Across episodes,
physical feedback is used to revise historical knowledge, update its
applicability, and organize reusable entries for subsequent retrieval.
Experiments on RMBench show that HPK improves average success by up to
24.2 percentage points across different agent models. With 80 interaction
rollouts, held-out success rises from 48.3\% to 75.0\% for GPT-5.5 and from
70.0\% to 88.3\% for GPT-6. RoboHarn-Evo also resolves over 83\% of historical
knowledge errors while retaining 95.8\% of valid knowledge, and transfers
zero-shot from RMBench to RoboDojo with gains of 35.0 and 25.0 percentage
points. These results demonstrate that physical interaction can be accumulated
into reusable knowledge for improving subsequent manipulation.
\end{abstract}

\section{Introduction}
Recent embodied agents increasingly use execution feedback to adapt their behavior during physical interaction~\citep{li2026roboclaw,chen2026show}. Beyond correcting the current execution, recent systems retain interaction experience as persistent memories~\citep{zhang2026harness,jiangxskill}, physical knowledge~\citep{li2026learning}, reusable skills~\citep{ju2026embodiskill}, or evolving runtime components~\citep{ding2026zetta,wang2026shaper}, allowing past interactions to inform future behavior. Recent language agents also refine reusable skills from execution feedback~\citep{yang2026skillopt,wang2026skillx}. These advances point toward a broader goal: enabling robots to improve through physical experience without repeatedly retraining their underlying models. A central challenge, however, is how to turn accumulated experience into \textbf{\emph{reliable improvement in subsequent behavior}}.

Hierarchical skill representations support compositional generalization
in robotic policies~\citep{xie2026skillnet}, while language agents evolve
hierarchical skill libraries from execution experience
\citep{xia2026skillrl,wang2026skillx}. For manipulation, we study how physical feedback should revise knowledge at different decision levels, where a local action effect need not establish task completion. At the \emph{task level}, the agent must decide which subtask to execute and when its objective has been satisfied~\citep{shi2025hi}; at the \emph{action level}, it must determine the geometry and physical interaction needed to realize that subtask~\citep{huang2023voxposer,huang2024rekep}. As illustrated in Fig.~\ref{fig:teaser}(a), the robot may switch to the next block before confirming completion of the current subtask (top), or select the correct target but release the cover off target (bottom). A successful grasp in the latter example can support the grasping strategy, but does not establish that the cover has been placed at the intended location. This motivates evolving task and action knowledge using their respective completion and effect criteria, while preserving the task goal during reuse.

To address this challenge, we introduce \textbf{RoboHarn-Evo}, a harness that evolves \textbf{Hierarchical Physical Knowledge (HPK)} while keeping the base vision-language model (VLM) and low-level executor fixed. HPK organizes manipulation experience into \emph{Task Knowledge}, which captures what subtask to execute and its completion condition, and \emph{Action Knowledge}, which captures object-relative geometric strategies and their intended and observed physical effects. As shown in Fig.~\ref{fig:teaser}(b), RoboHarn-Evo couples knowledge use and knowledge evolution through an \emph{inner execution loop} and an \emph{outer knowledge-update loop}. The inner execution loop retrieves Task Knowledge for subtask planning and Action Knowledge for action grounding in the current scene. The outer knowledge-update loop uses recorded observations, tool calls, and physical outcomes to extract and consolidate individual pieces of task or action guidance (\emph{knowledge entries}). It checks Task Knowledge against subtask completion and Action Knowledge against intended physical effects. These checks guide content revision and determine eligibility for subsequent retrieval. By linking subtasks to their constituent operations, HPK preserves the relationship between task context and local physical experience for future reuse.

We assess \textbf{\emph{self-improvement}} through two complementary criteria with the VLM and executor held fixed. First, \emph{knowledge correction}: accumulated physical evidence should improve persistent knowledge by correcting or deactivating erroneous entries while preserving valid ones. Second, \emph{held-out behavioral improvement}: the resulting knowledge should improve performance on held-out configurations when the knowledge store is frozen and evaluation feedback is excluded from updates. We additionally evaluate \emph{cross-domain knowledge transfer} to examine whether learned knowledge generalizes beyond the environment in which it was acquired.

We evaluate RoboHarn-Evo on RMBench~\citep{chen2026rmbench} and RoboDojo~\citep{chen2026robodojo} using different VLM backbones and knowledge configurations, with policy-based methods and Harness VLA as reference systems. On RMBench, full HPK improves average success by up to 24.2 percentage points over the same harness without HPK, and ablations support the complementary benefits of Task Knowledge and Action Knowledge. As shown in Fig.~\ref{fig:teaser}(c), after 80 interaction rollouts, mean held-out success increases from 48.3\% to 75.0\% with GPT-5.5 and from 70.0\% to 88.3\% with GPT-6. An external GPT-6 audit finds that RoboHarn-Evo repairs or deactivates 10 of the 12 initially incorrect knowledge entries (83.3\%) with GPT-5.5 and 11 of 12 (91.7\%) with GPT-6. Among the separate set of 24 initially correct entries, both models retain 23 (95.8\%). In the additional generalization evaluation, transferring frozen HPK zero-shot from RMBench to RoboDojo improves average success by 35.0 and 25.0 percentage points with GPT-5.5 and GPT-6, respectively. These results support the use of evolving hierarchical physical knowledge to improve subsequent manipulation under a fixed base model and executor.

Our contributions are threefold. 
\textbf{(1) Experience-driven self-improvement for robotic manipulation.} We formulate robotic self-improvement as the continual refinement and reuse of physical experience under a fixed base VLM and executor, and introduce RoboHarn-Evo to couple online execution with persistent knowledge evolution.
\textbf{(2) Evidence-driven evolution of HPK.} We link Task Knowledge for subtask selection and completion with Action Knowledge for object-relative geometry and physical effects. Level-specific verification guides knowledge revision and retrieval eligibility, while task-goal consistency constrains the reuse of geometric strategies.
\textbf{(3) Systematic evaluation of knowledge evolution.} Experiments on RMBench and RoboDojo show complementary benefits from the two knowledge levels, correction of erroneous historical knowledge while retaining valid knowledge, continued improvement on held-out configurations, and zero-shot cross-domain transfer.

\section{Related Work}
\label{sec:related_work}

\paragraph{Harnesses for agentic manipulation.}
Agentic manipulation wraps robot policies in a reasoning-and-execution loop
that selects capabilities, monitors outcomes, and replans from feedback.
SayCan and Inner Monologue established affordance-grounded skill selection
and feedback-driven replanning
\citep{ichter2023saycan,huang2023innermonologue}.
Recent systems extend this paradigm to learned robot policies:
VLAs-as-Tools and Goal2Skill couple high-level VLM reasoning with VLA
execution for long-horizon manipulation
\citep{lei2026vlastools,liu2026goal2skill}, while RoboHarness and Harness VLA
augment frozen VLAs with memory-guided orchestration and external
interventions or analytic primitives
\citep{li2026roboharness,zhang2026harness}.
A systematic study further identifies planner--executor orchestration as a
key determinant of hierarchical VLA performance
\citep{hu2026orchestrating}.
SkillNet models hierarchical skills within a learned VLA~\citep{xie2026skillnet}. RoboHarn-Evo evolves physical knowledge reused by a fixed planner--executor system.

\paragraph{Self-evolution in robotics.}
SkillOpt refines external skills from execution feedback~\citep{yang2026skillopt}; XSkill consolidates task-level skills and action-level experiences without parameter updates~\citep{jiangxskill}. Hierarchical skill evolution is also studied by SkillRL, SkillPyramid, and SPyCE~\citep{xia2026skillrl,xiong2026skillpyramid,zhang2026spyce}. In robotics, REFLECT and AIC support failure correction and geometric adjustment, while PhysMem verifies interaction-derived physical knowledge~\citep{liu2023reflect,xiong2025aic,li2026physmem}. Uni-Skill extends robot skills through a hierarchical demonstration repository~\citep{xie2026uniskill}. EmbodiSkill, SHAPER, Zetta, and ASPIRE evolve procedural skills, runtime components, or executable code~\citep{ju2026embodiskill,wang2026shaper,ding2026zetta,lu2026aspire}. Other systems refine policies through autonomous data collection~\citep{li2026roboclaw,xiao2026enpire}. RoboHarn-Evo focuses on physical knowledge evolution under a fixed VLM and executor: it checks subtask completion separately from local action effects to guide knowledge revision and retrieval eligibility, while preserving task goals during reuse.

\section{Problem Formulation}
\label{sec:problem_formulation}

We study experience-driven self-improvement in robotic manipulation, where physical experience continually refines persistent knowledge under a fixed VLM and executor.
Let $\mathcal K_i$ denote the knowledge available before episode $i$.
For instruction $l$, episode $i$ is recorded at operation boundaries as
$\tau_i=(l;\,o_0,a_0,o_1,a_1,\ldots,o_{T_i})$,
where $o_t$ is the observation and $a_t$ the robot-action sequence for operation $t$.
With runtime memory $s_t$, retrieved knowledge conditions the VLM's tool-use decisions:
\begin{equation}
c_t\sim p_\theta
\!\left(\cdot\mid s_t,l,\mathcal K_i\right),
\qquad
a_t=\operatorname{Exec}(s_t,c_t),
\label{eq:tool-execution}
\end{equation}
where $c_t$ is a parameterized tool call and the executor grounds it in the current scene with execution feedback.

Across episodes, a knowledge-update rule $\mathcal U$ revises the store using observed task progress and physical action effects:
\begin{equation}
\mathcal K_{i+1}
=
\mathcal U\!\left(
\mathcal K_i,
\operatorname{View}(\tau_i)
\right),
\label{eq:knowledge-evolution}
\end{equation}
where $\operatorname{View}(\tau_i)$ contains observations, tool calls, and execution feedback.
Each episode is executed with $\mathcal K_i$ and updates the knowledge used in subsequent episodes.
We seek an update rule that yields higher task success as interaction accumulates, evaluated using frozen knowledge snapshots on fixed held-out configurations without feeding evaluation outcomes back into the store.

\begin{figure*}[t]
\centering
\includegraphics[width=\textwidth]{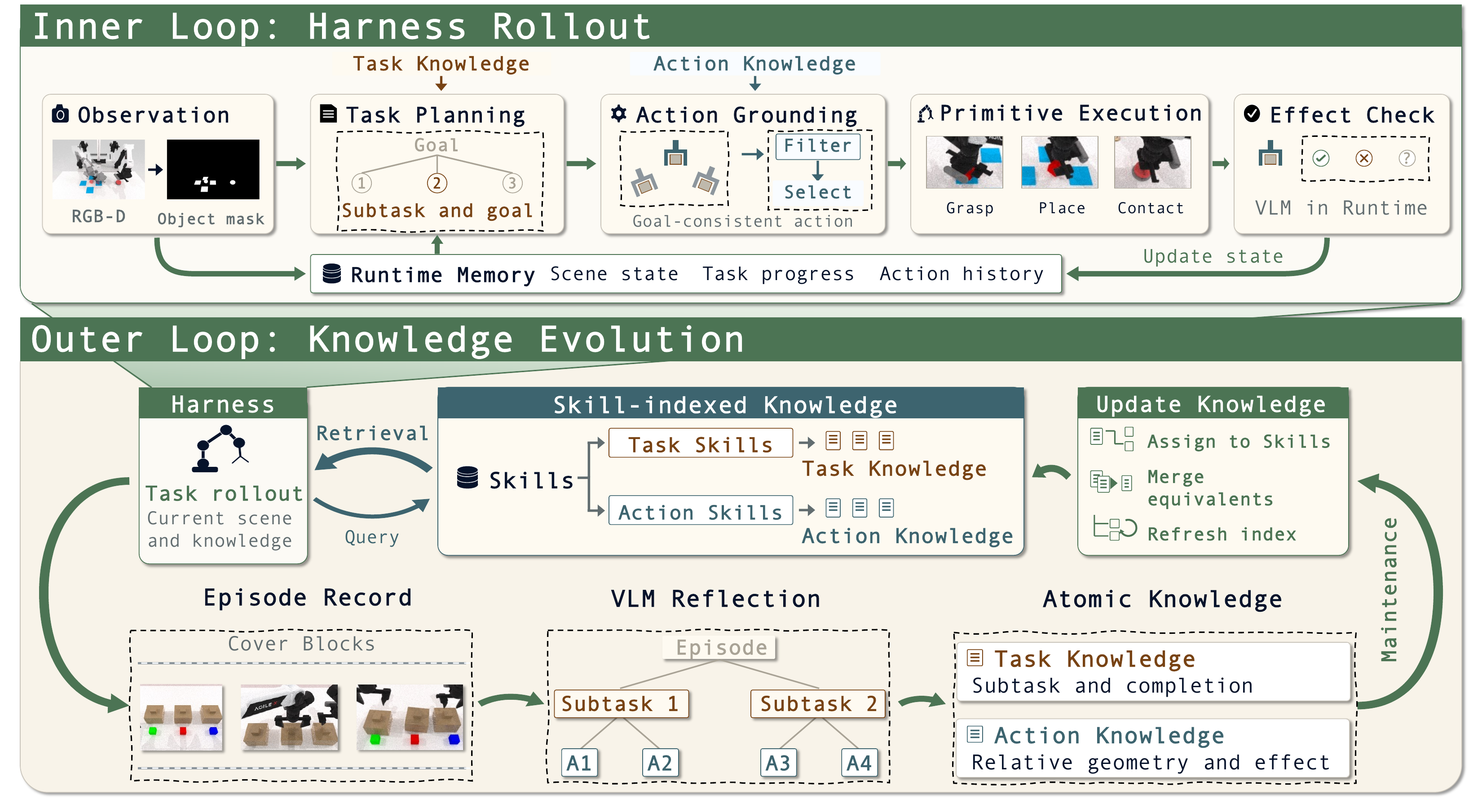}
\caption{\textbf{Overview of RoboHarn-Evo.} \textbf{Top: }The inner loop executes manipulation operations and records their outcomes. \textbf{Bottom: }The outer loop organizes this experience into Task and Action Knowledge for subsequent decisions. Concrete motions are computed from the current scene.}
\label{fig:Method_overview}
\end{figure*}

\section{Method}
\label{sec:method}

RoboHarn-Evo connects a fixed VLM with robot execution through two loops. The inner loop uses \emph{Hierarchical Physical Knowledge} (HPK) to guide manipulation (Sec.~\ref{sec:method_backbone}); the outer loop extracts and maintains this knowledge from observed physical outcomes (Sec.~\ref{sec:method_knowledge} and \ref{sec:method_maintenance}).

\subsection{Agentic Manipulation Harness}
\label{sec:method_backbone}

As shown in the top of Fig.~\ref{fig:Method_overview}, the inner loop maintains runtime memory $s_t$ from RGB-D observations, object segmentation, task progress, and action history. Let
$\mathcal K_i=(\mathcal K_i^{\mathrm T},\mathcal K_i^{\mathrm A})$
denote the Task and Action Knowledge available before episode $i$. The VLM first chooses a task strategy $u_t$ and then an action strategy $z_t$, which together condition its tool call:
\begin{equation}
\begin{aligned}
u_t
\sim p_\theta
\!\left(\cdot\mid s_t,l,\mathcal K_i^{\mathrm T}\right),
z_t
\sim p_\theta
\!\left(\cdot\mid s_t,u_t,\mathcal K_i^{\mathrm A}\right),
c_t
\sim p_\theta
\!\left(\cdot\mid s_t,u_t,z_t\right).
\end{aligned}
\label{eq:policy-factorization}
\end{equation}
Here, $u_t$ specifies the current subtask, target relation, and completion condition, while $z_t$ specifies object-relative geometry for an operation
$\alpha_t\in\{\texttt{grasp},\texttt{place},\texttt{contact}\}$.
The harness grounds $c_t$ in the current scene, and the robot executor produces $a_t$ through Eq.~\afkeqref{eq:tool-execution}.

After each operation, the effect check records the observed physical outcome and updates $s_t$. Local action effects and subtask completion are evaluated separately: an operation may succeed physically without completing the enclosing subtask. Verified completion advances task planning; otherwise the updated runtime state conditions the next decision. The resulting observations, tool calls, and effect checks form the episode record used by the outer loop.

\subsection{Hierarchical Physical Knowledge}
\label{sec:method_knowledge}

\paragraph{From trajectories to atomic knowledge.}
The bottom of Fig.~\ref{fig:Method_overview} shows how an episode is converted into HPK. A VLM reflector $R_\phi$ maps the observable trajectory into an ordered hierarchy of Task and Action Knowledge:
\begin{equation}
\mathcal P_i
=
R_\phi\!\left(\operatorname{View}(\tau_i)\right)
=
\left(
k_{i,j}^{\mathrm T},
\left(k_{i,j,r}^{\mathrm A}\right)_{r=1}^{m_{i,j}}
\right)_{j=1}^{m_i}.
\label{eq:afk-package}
\end{equation}
Each Task entry $k_{i,j}^{\mathrm T}$ captures when a subtask applies, what it should achieve, and when it is complete. Its associated Action entries $k_{i,j,r}^{\mathrm A}$ capture the operation type, object-relative geometric strategy, and observed physical effect. Linking each subtask to the operations that realize it preserves task context while keeping local action success distinct from task completion.

\paragraph{From atomic knowledge to skill-indexed knowledge.}
The extracted atomic entries are integrated into the persistent stores
$\mathcal K_{i+1}^{\mathrm T}$ and $\mathcal K_{i+1}^{\mathrm A}$ and organized by a Skill index
$\mathcal F_{i+1}=(\mathcal F_{i+1}^{\mathrm T},\mathcal F_{i+1}^{\mathrm A})$, as illustrated in the bottom of Fig.~\ref{fig:Method_overview}. Each Skill is a retrieval index that groups contextually related entries and summarizes when they are useful; execution is conditioned on the selected atomic Task or Action Knowledge. The task--action schema and execution primitives remain fixed; physical experience evolves knowledge content, applicability, and retrieval eligibility. Evidence representation and update details are provided in Appendix~\ref{sec:app_method_reflection_implementation}.

\begin{figure*}[t]
\centering
\includegraphics[width=\textwidth]{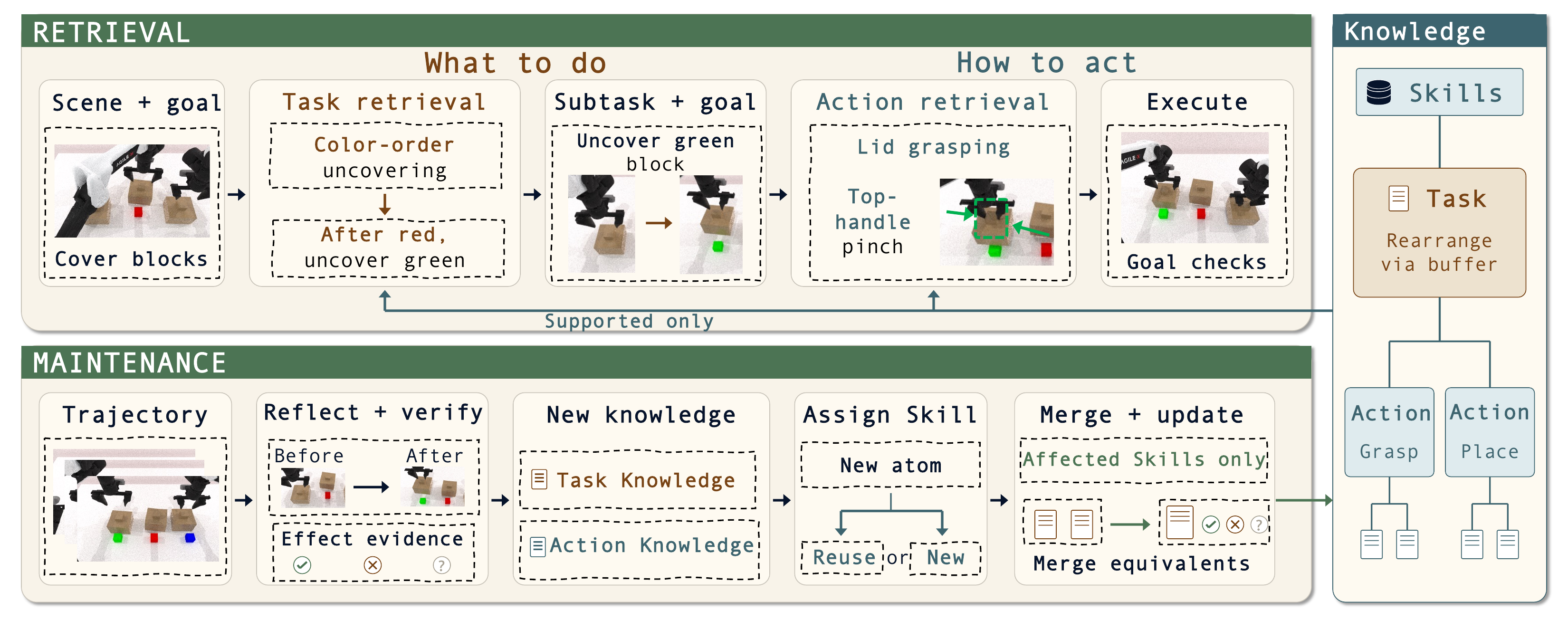}
\caption{\textbf{Skill-routed retrieval and evidence-driven maintenance.}
\textbf{Top:} Task retrieval determines \emph{what to do}; conditioned on that subtask, Action retrieval determines \emph{how to act} using supported atomic knowledge.
\textbf{Bottom:} Before--after execution evidence is reflected into new Task and Action Knowledge, assigned to Skills, and used to consolidate and update only the affected knowledge groups.}
\label{fig:afk_retrieval_maintenance}
\end{figure*}

\subsection{Knowledge retrieval and maintenance}
\label{sec:method_retrieval}
\label{sec:method_maintenance}

\paragraph{Retrieval.}
The top of Fig.~\ref{fig:afk_retrieval_maintenance} mirrors the two decisions in Eq.~\ref{eq:policy-factorization}. Given the scene, goal, and runtime memory, Task retrieval selects applicable knowledge for $u_t$, answering \emph{what to do} next. Conditioned on that subtask and primitive $\alpha_t$, Action retrieval selects knowledge for $z_t$, answering \emph{how to act}. Skill summaries route each query to relevant groups, after which the VLM selects supported atomic entries whose conditions match the current context. The retrieved task strategy constrains action grounding so that the selected geometry remains consistent with the intended object relation and subtask goal.

\paragraph{Maintenance.}
The bottom of Fig.~\ref{fig:afk_retrieval_maintenance} shows how each rollout updates the persistent knowledge used by later episodes:
\begin{equation}
(\mathcal K_{i+1},\mathcal F_{i+1})
=
\operatorname{Maintain}(\mathcal K_i,\mathcal F_i,\mathcal P_i).
\label{eq:afk-maintenance}
\end{equation}
Verification follows the HPK hierarchy. Task Knowledge is assessed against subtask completion, whereas Action Knowledge is assessed against its intended physical effect. Thus, a successful grasp can support an Action entry even when the enclosing placement subtask remains incomplete.

For knowledge that is applicable and faithfully executed, the observed outcome provides \texttt{support}, \texttt{oppose}, or \texttt{unverified} evidence. Maintenance uses this evidence to merge equivalent entries and revise conditions or strategies challenged by subsequent interaction. Revised content becomes retrievable only when supported without applicable opposition, and only the affected Skill summaries are refreshed. Detailed evidence attribution, versioning, and consolidation procedures are given in Appendix~\ref{sec:app_method_reflection_implementation}.

\providecommand{\FloatBarrier}{}
\providecommand{\AFKQFigureDir}{figures}

\begin{table}[!t]
\caption{\textbf{Task success on RMBench.}
We report task success rates (\%) on six RMBench tasks, with Overall denoting
the six-task mean. X-VLA and Mem-0 results are reported from the Mem-0
benchmark evaluation; all remaining results are evaluated on 20 held-out
episodes per task. Within each model, w/o HPK and Full HPK share the same
executor. HarnessVLA is tested with GPT-5.5.}
\label{tab:rmbench_main}

\centering
\fontsize{7.4}{8.5}\selectfont
\setlength{\tabcolsep}{1.15pt}
\renewcommand{\arraystretch}{1.12}

\begin{tabular*}{\linewidth}{
@{\extracolsep{\fill}}
l
c c c c
c c
c c
c c
@{}
}
\toprule

\multicolumn{1}{c}{\multirow{2}{*}[-0.4ex]{Task}}
& \multirow{2}{*}[-0.4ex]{$\pi_{0.5}$}
& \multirow{2}{*}[-0.4ex]{X-VLA}
& \multirow{2}{*}[-0.4ex]{Mem-0}
& \multirow{2}{*}[-0.4ex]{HarnessVLA}
& \multicolumn{2}{c}{Qwen3.8-27B}
& \multicolumn{2}{c}{GPT-5.5}
& \multicolumn{2}{c}{GPT-6}
\\

\cmidrule(lr){6-7}
\cmidrule(lr){8-9}
\cmidrule(l){10-11}

& & & & &
w/o HPK & \textbf{Full HPK}
& w/o HPK & \textbf{Full HPK}
& w/o HPK & \textbf{Full HPK}
\\

\midrule

Rearrange Blocks
& 13\% & 13\% & \underline{89\%} & 30\%
& 15\% & 35\%
& 50\% & 80\%
& 75\% & \textbf{90\%}
\\

Swap Blocks
& 24\% & 16\% & 67\% & 35\%
& 10\% & 25\%
& 45\% & \underline{70\%}
& \underline{70\%} & \textbf{85\%}
\\

Press Button
& 0\% & 0\% & 0\% & 75\%
& 35\% & 60\%
& \underline{95\%} & \textbf{100\%}
& \textbf{100\%} & \textbf{100\%}
\\

Swap T
& 15\% & 3\% & 14\% & 35\%
& 20\% & 40\%
& 55\% & \underline{90\%}
& 45\% & \textbf{95\%}
\\

Put Back Block
& 11\% & 18\% & \textbf{90\%} & 25\%
& 20\% & 50\%
& 50\% & \underline{75\%}
& \underline{75\%} & \textbf{90\%}
\\

Cover Blocks
& 0\% & 2\% & 68\% & 50\%
& 15\% & 40\%
& 55\% & \underline{80\%}
& 70\% & \textbf{90\%}
\\

\midrule

\textbf{Overall}
& 10.5\%
& 8.7\%
& 54.7\%
& 41.7\%
& 19.2\%
& 41.7\%
& 58.3\%
& \underline{82.5\%}
& 72.5\%
& \textbf{91.7\%}
\\

\bottomrule
\end{tabular*}
\end{table}

\section{Experiments}
\label{sec:experiments}

In this section, we aim to address the following four research questions: \textbf{Q1:} Does \emph{Hierarchical Physical Knowledge} improve manipulation, and how do \emph{Task} and \emph{Action Knowledge} contribute?\textbf{Q2:} Can \emph{RoboHarn-Evo} resolve historical errors while preserving valid knowledge? \textbf{Q3:} Does continued interaction improve performance on held-out configurations? \textbf{Q4:} Can source knowledge support execution across benchmarks and sim2real environments?

\paragraph{Experimental setup.} We evaluate \emph{RoboHarn-Evo} on six RMBench~\citep{chen2026rmbench} tasks and assess cross-benchmark knowledge transfer on RoboDojo~\citep{chen2026robodojo}. The main comparison uses Qwen3.8-27B, GPT-5.5, and GPT-6, with $\pi_{0.5}$~\citep{physicalintelligence2025pi05} and Harness VLA~\citep{zhang2026harness} as reference methods. Detailed tasks, comparisons, and protocols appear in Appendix~\ref{sec:exp_setup}.

\FloatBarrier
\subsection{Effectiveness of Hierarchical Physical Knowledge}
\label{sec:exp_hierarchy}
\label{sec:exp_main_rmbench}

\textbf{Comparative Performance.} As shown in Tab.~\ref{tab:rmbench_main}, averaged across the three VLM backbones, RoboHarn-Evo exceeds $\pi_{0.5}$ and Harness VLA by 61.1 and 30.3 percentage points in overall task success, respectively. \textbf{Gains across VLM Backbones.}
Tab.~\ref{tab:rmbench_main} compares w/o HPK and w/ Full HPK across six tasks. Full HPK raises mean success from 19.2\% to 41.7\% with Qwen3.8-27B, from 58.3\% to 82.5\% with GPT-5.5, and from 72.5\% to 91.7\% with GPT-6. Gains span task planning and action grounding: with GPT-5.5, success increases from 50.0\% to 80.0\% on Rearrange Blocks and from 55.0\% to 90.0\% on Swap T, respectively.

\begin{minipage}{\linewidth}
\newsavebox{\HPKCompactAblation}
\sbox{\HPKCompactAblation}{%
  \footnotesize
  \setlength{\tabcolsep}{4pt}%
  \renewcommand{\arraystretch}{1.02}%
  \begin{tabular}{@{}lrr@{\hspace{12pt}}rr@{}}
  \toprule
  \multirow{2}{*}{Method}
  & \multicolumn{2}{c}{Qwen3.8-27B}
  & \multicolumn{2}{c}{GPT-5.5} \\
  \cmidrule(lr){2-3}\cmidrule(l){4-5}
  & Overall (\%) & $\Delta$
  & Overall (\%) & $\Delta$ \\
  \midrule
  w/o HPK & 19.2 & 0.0 & 58.3 & 0.0 \\
  Flat Reflection & 24.2 & +5.0 & 64.2 & +5.8 \\
  Task-only & 27.5 & +8.3 & \underline{70.8} & \underline{+12.5} \\
  Action-only & \underline{28.3} & \underline{+9.2} & 66.7 & +8.3 \\
  \textbf{RoboHarn-Evo}
  & \textbf{41.7} & \textbf{+22.5}
  & \textbf{82.5} & \textbf{+24.2} \\
  \bottomrule
  \end{tabular}%
}

\setlength{\intextsep}{0pt}
\setlength{\columnsep}{12pt}
\begin{wraptable}{r}{\wd\HPKCompactAblation}
\vspace{-1.5pt} 
\captionsetup{font=normalsize,justification=raggedright,
  singlelinecheck=false,width=\linewidth,position=top,skip=4pt}
\caption{\textbf{Component ablation on RMBench.}
$\Delta$ is the overall gain over w/o HPK in percentage points.}
\label{tab:afk_ablation}
\usebox{\HPKCompactAblation}
\par\vspace{8pt} 
\end{wraptable}
\paragraph{Complementarity of Knowledge Levels.}
  Tab.~\ref{tab:afk_ablation} evaluates the individual and combined
  contributions of Task and Action Knowledge using the same source
  experience. With Qwen3.8-27B, Task-only and Action-only achieve
  27.5\% and 28.3\% overall success, respectively, compared with
  19.2\% without HPK. With GPT-5.5, they achieve 70.8\% and 66.7\%,
  compared with 58.3\% without HPK. Combining both levels raises
  success to 41.7\% and 82.5\%, respectively, outperforming either
  level alone and supporting their complementary contributions.
  Full HPK also exceeds Flat Reflection, which achieves 24.2\%
  and 64.2\%, indicating the benefit of organizing and applying
  the same experience at distinct decision levels.
\par
\ifnum\value{WF@wrappedlines}>1
  \vspace{\dimexpr\value{WF@wrappedlines}\baselineskip-\baselineskip\relax}
\fi
\WFclear
\end{minipage}

\begin{figure}[!t]
\centering
\includegraphics[width=\linewidth]{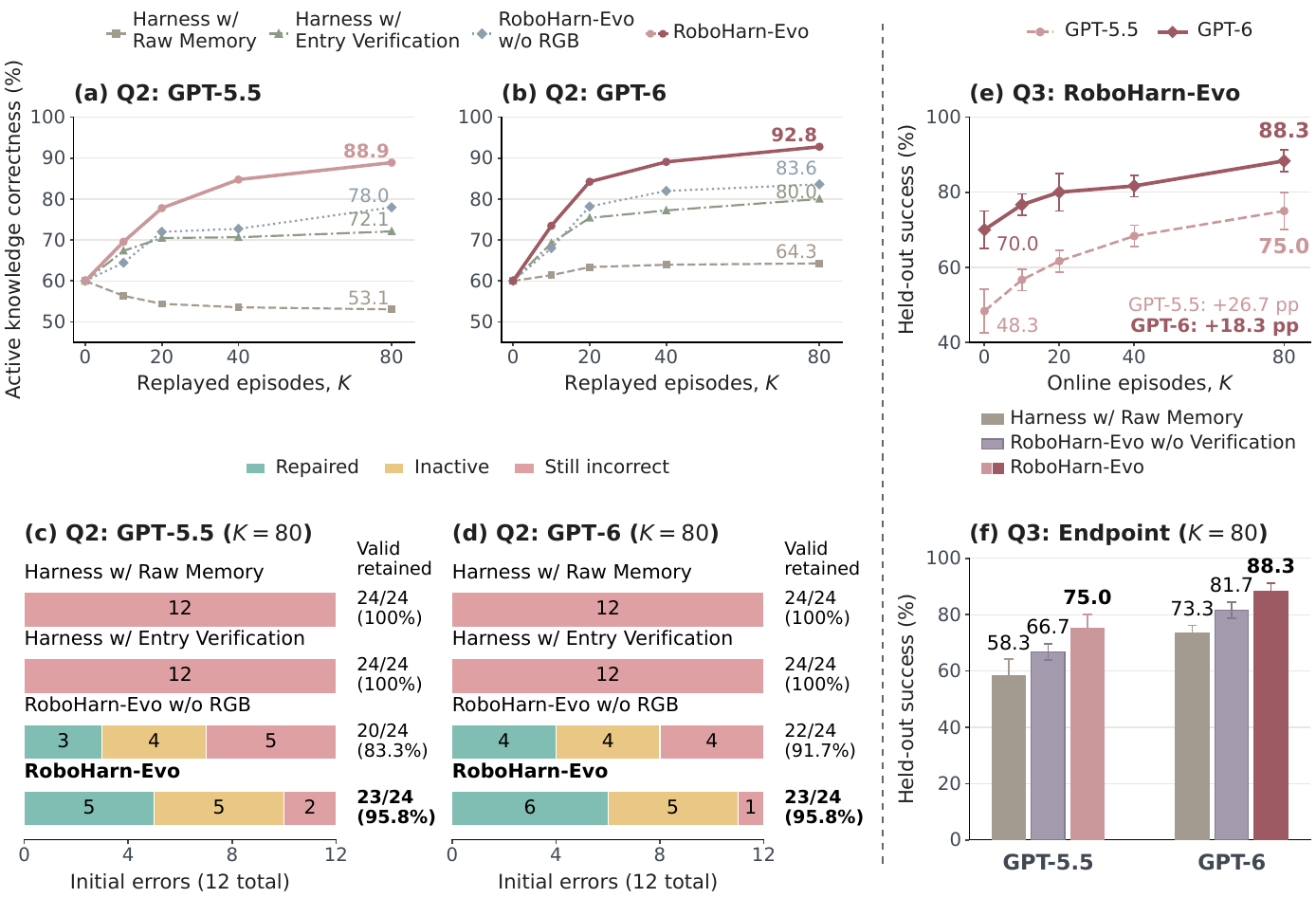}
\caption{\textbf{Knowledge maintenance and held-out manipulation performance.} \textbf{(a,b)} Active-knowledge correctness under shared-evidence replay. \textbf{(c,d)} Outcomes of the 12 initial errors and retention of the 24 initially correct entries at $K=80$. \textbf{(e)} Held-out success across learning checkpoints. \textbf{(f)} Endpoint comparisons using the same source experience at $K=80$. $K$ counts interaction rollouts. Error bars in (e,f) indicate standard deviations over three learning histories. Q2 replays previously collected interaction records to evaluate knowledge maintenance; Q3 collects experience through online interaction to evaluate subsequent task performance.}
\label{fig:knowledge_and_capability_growth}
\label{fig:self_evolution_curve}
\end{figure}

\FloatBarrier
\subsection{Knowledge Maintenance}
\label{sec:exp_self_correction}

We evaluate knowledge maintenance on \emph{Swap Blocks} and
\emph{Put Back Block} from RMBench. For each model, four variants start from the same historical pool of 40 entries (24 correct, 12 incorrect, and 4 unverified) and process identical evidence from one predesignated interaction history. \emph{Harness w/ Raw Memory} only appends entries; \emph{Harness w/ Entry Verification} checks only additions; and \emph{RoboHarn-Evo w/o RGB} reviews historical knowledge using textual evidence. Full \emph{RoboHarn-Evo} additionally incorporates visual evidence. Following the LLM-as-a-judge paradigm~\cite{li2025generation}, knowledge correctness is assessed by an external GPT-6 judge, separate from RoboHarn-Evo's execution and knowledge maintenance components, using recorded observations and execution evidence.

Fig.~\ref{fig:knowledge_and_capability_growth}(a,b) tracks the
overall quality of the active knowledge pool, measured as the
proportion of active entries judged correct. At $K=80$,
RoboHarn-Evo reaches 88.9\% with GPT-5.5 and 92.8\% with GPT-6.
To distinguish historical error correction from changes in pool
composition, Fig.~\ref{fig:knowledge_and_capability_growth}(c,d) track the fixed cohorts of initially
incorrect and correct entries. At the same checkpoint,
RoboHarn-Evo repairs 41.7\% of the initial errors with GPT-5.5
and 50.0\% with GPT-6, while deactivating another 41.7\% with
either model. Both retain 95.8\% of initially correct knowledge.
Raw Memory and Entry Verification leave all initial errors
unresolved, while removing RGB reduces both the repair rate
and valid-knowledge retention.

 \subsection{Self-Improvement through Continued Interactions}
\label{sec:exp_self_evolution}

We evaluate whether accumulated experience improves manipulation
on held-out configurations of \emph{Swap Blocks} and
\emph{Put Back Block}. For each model, RoboHarn-Evo completes
three independent learning histories of 80 interaction rollouts,
each starting from an empty knowledge store. Model parameters
and execution tools remain fixed throughout. At predefined
checkpoints, knowledge snapshots are frozen and evaluated on
the same 20 held-out configurations. Evaluation feedback is
excluded from knowledge updates.

For each learning history, we replay its 80 source trajectories
to construct two endpoint control stores.
\emph{Harness w/ Raw Memory} retains raw experience, while
\emph{RoboHarn-Evo w/o Verification} constructs knowledge without
physical-evidence verification. Both controls are evaluated
only at $K=80$, using the same held-out configurations and
evaluation protocol as RoboHarn-Evo. Runtime execution checks
remain enabled in all variants.

Fig.~\ref{fig:knowledge_and_capability_growth} (e) shows that mean
held-out success increases from 48.3\% to 75.0\% with GPT-5.5 and
from 70.0\% to 88.3\% with GPT-6 as interaction accumulates. Under shared source experience at $K=80$, Fig.~\ref{fig:knowledge_and_capability_growth} (f) shows that RoboHarn-Evo exceeds Raw Memory by 16.7 and 15.0 percentage points for GPT-5.5 and GPT-6, respectively. It also exceeds w/o Verification by 8.3 and 6.7 percentage points. These endpoint comparisons support the value of constructing and verifying reusable knowledge from the same interaction evidence.

\subsection{Knowledge Transfer across Environments}
\label{sec:exp_transfer}

\begin{table}[!t]
\caption{\textbf{Zero-shot knowledge transfer from RMBench to RoboDojo.}
Score and success rate (SR) are reported as percentages, with 10 target
episodes per task. $\Delta$ is the difference in Average between
\mbox{w/ Full HPK} and \mbox{w/o HPK}, in percentage points.}
\label{tab:transfer_summary}
\centering
\footnotesize
\setlength{\tabcolsep}{3.2pt}
\renewcommand{\arraystretch}{1.12}
\begin{tabular*}{\linewidth}{@{\extracolsep{\fill}}l*{8}{c}@{}}
\toprule
\multicolumn{1}{c}{\multirow{3}{*}[-0.8ex]{Task}}
& \multicolumn{4}{c}{GPT-5.5}
& \multicolumn{4}{c}{GPT-6} \\
\cmidrule(lr){2-5}\cmidrule(l){6-9}
& \multicolumn{2}{c}{w/o HPK}
& \multicolumn{2}{c}{\textbf{w/ Full HPK}}
& \multicolumn{2}{c}{w/o HPK}
& \multicolumn{2}{c}{\textbf{w/ Full HPK}} \\
\cmidrule(lr){2-3}\cmidrule(lr){4-5}
\cmidrule(lr){6-7}\cmidrule(l){8-9}
& Score & SR & Score & SR & Score & SR & Score & SR \\
\midrule
Cover Blocks
& 37.0 & 30.0 & \textbf{75.0} & \textbf{70.0}
& 57.0 & 50.0 & \textbf{93.0} & \textbf{90.0} \\
Press by Number
& 50.0 & 50.0 & \textbf{80.0} & \textbf{80.0}
& 80.0 & 80.0 & \textbf{90.0} & \textbf{90.0} \\
\midrule
\textbf{Average}
& 43.5 & 40.0 & \textbf{77.5} & \textbf{75.0}
& 68.5 & 65.0 & \textbf{91.5} & \textbf{90.0} \\
$\Delta$ (pp)
& \multicolumn{2}{c}{--} & \textbf{+34.0} & \textbf{+35.0}
& \multicolumn{2}{c}{--} & \textbf{+23.0} & \textbf{+25.0} \\
\bottomrule
\end{tabular*}
\end{table}

\begin{figure}[!t]
\centering

\setbox0=\hbox{%
  \includegraphics[
    width=0.65\linewidth,
    pagebox=cropbox
  ]{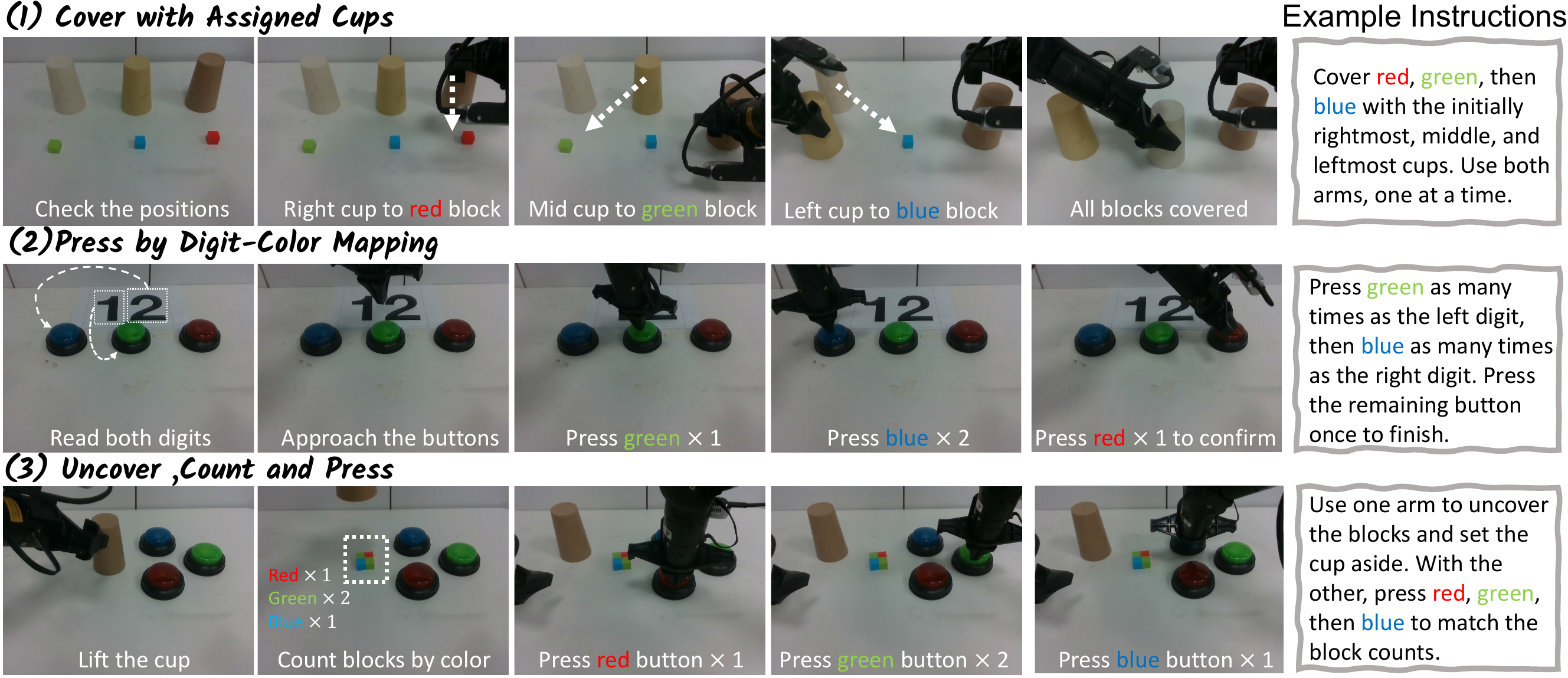}%
}
\edef\RealPanelHeight{\the\dimexpr\ht0+\dp0\relax}

\begin{minipage}[t]{0.65\linewidth}
  \vspace{0pt}
  \centering
  {\small\textbf{(a) Real-world manipulation tasks}\par}
  \vspace{0.35em}
  \includegraphics[
    height=\RealPanelHeight,
    pagebox=cropbox
  ]{Figure/real.pdf}
\end{minipage}\hfill
\begin{minipage}[t]{0.34\linewidth}
  \vspace{0pt}
  \centering
  {\small\textbf{(b) Real-world performance}\par}
  \vspace{0.35em}
  \includegraphics[
    height=\RealPanelHeight
  ]{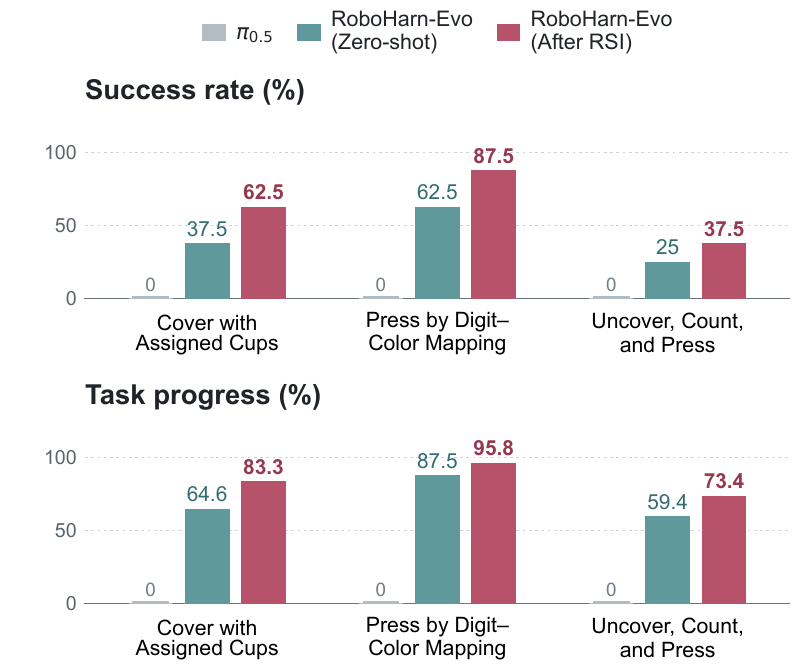}
\end{minipage}

\caption{\textbf{Real-world transfer and self-improvement.}
\textbf{(a)} Task sequences and example instructions.
\textbf{(b)} Success rate \textbf{(top)} and normalized task progress
\textbf{(bottom)} for $\pi_{0.5}$ and RoboHarn-Evo before and after RSI.
Zero-shot uses simulation-derived HPK; After RSI incorporates
additional real-world interaction.
RoboHarn-Evo results use eight scenes per task, with full task
completion counted as success.}
\label{fig:real-robot-tasks}
\end{figure}

\paragraph{Cross-benchmark transfer.}
We transfer HPK acquired on RMBench to \emph{Cover Blocks} and
\emph{Press by Number} in RoboDojo. The source knowledge remains
frozen throughout evaluation, while tool arguments and motion
targets are determined from current observations. The \emph{w/o HPK}
and \emph{w/ Full HPK} conditions share the target agent, executor,
initial configurations, and action budget.

Tab.~\ref{tab:transfer_summary} shows that transferred HPK raises
mean success from 40.0\% to 75.0\% with GPT-5.5 and from 65.0\%
to 90.0\% with GPT-6. Mean task-progress scores also increase
from 43.5 to 77.5 and from 68.5 to 91.5, respectively.
Both models improve on both target tasks without updating
knowledge in the target environment.

\paragraph{Self-improvement across real-world scenes.}
We deploy RoboHarn-Evo on an AgileX Cobot Magic platform for the three
tasks in Fig.~\ref{fig:real-robot-tasks}(a).
We compare $\pi_{0.5}$ with RoboHarn-Evo using simulation-derived HPK
\emph{(Zero-shot)} and after further knowledge updates from real-world
interaction \emph{(After RSI)}. The VLM and executor remain fixed during
these updates. Across eight scenes per task,
Fig.~\ref{fig:real-robot-tasks}(b) shows that RSI raises mean success
from 41.7\% to 62.5\% and mean normalized task progress from 70.5\%
to 84.2\%, averaged across the three tasks. Both metrics improve on
every task. These results support the use of simulation-derived
knowledge in real-world manipulation and its further improvement
through physical interaction. Hardware, task definitions, and scoring
rules are detailed in Appendix~\ref{sec:app_real_world_setting}.

\begin{figure*}[!t]
\centering
\includegraphics[width=\textwidth]{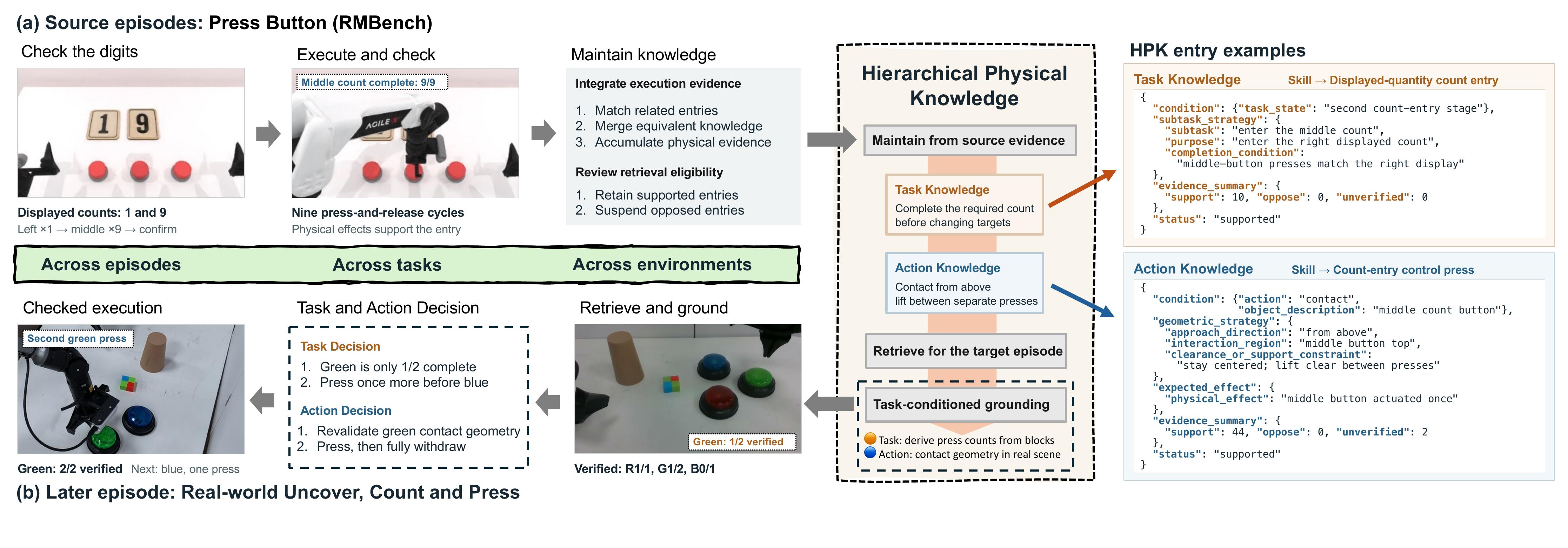}
\caption{\textbf{Task-conditioned knowledge reuse from simulation
to the real world.}
\textbf{(a)} Simulated pressing experience provides physical evidence
for maintaining HPK.
\textbf{(b)} A later real-world episode uses block-derived counts
and scene-specific contact geometry to complete the second green
press before switching to blue.
The right panels show condensed Task and Action Knowledge entries
with accumulated source evidence.
Arrows illustrate the maintenance and reuse workflow.}
\label{fig:Case_Study1}
\end{figure*}

\paragraph{Case study: task-conditioned knowledge reuse.}
Fig.~\ref{fig:Case_Study1} illustrates knowledge reuse from simulated
\emph{Press Button} to real-world \emph{Uncover, Count and Press}.
Source HPK captures guidance for completing the required count before
switching targets and performing distinct press-and-release cycles.
In the target task, the instruction and observed block counts determine
the button identities and required counts, while contact geometry is
recomputed from the current scene. With only one of two required green
presses verified, the robot selects another green press, revalidates
its contact geometry, and verifies the second press before selecting
blue. HPK remains fixed during this target episode.

\section{Conclusion}
\label{sec:conclusion}

We presented RoboHarn-Evo, a harness for experience-driven
self-improvement in robotic manipulation with a fixed VLM and
low-level executor. Its Hierarchical Physical Knowledge links
subtask decisions to object-relative action strategies, allowing
past experience to guide planning and execution under a shared
task goal. Physical feedback then supports the revision and
reuse of this knowledge across episodes. Experiments show that
Task and Action Knowledge contribute complementary benefits,
while evidence-driven maintenance repairs or deactivates historical
errors and preserves valid knowledge. Evaluation with frozen
knowledge snapshots demonstrates improved held-out performance
as interaction accumulates. Cross-benchmark and real-robot results
further show that simulation-derived knowledge supports transfer
and can be refined through subsequent physical interaction.
These findings support hierarchical knowledge evolution as a
mechanism for turning accumulated experience into improved
manipulation without updating model parameters.

\clearpage

\bibliography{references}
\bibliographystyle{iclr2027_conference}

\clearpage
\appendix
\numberwithin{equation}{section}
\section{Experimental Setup}

\subsection{Simulation Experimental setup}
\label{sec:exp_setup}

\paragraph{Benchmarks and tasks.}
RMBench~\citep{chen2026rmbench} is our main platform. We evaluate six tasks covering task planning, action grounding, and their interaction, with 20 held-out episodes per task. Knowledge correction and self-evolution use \emph{Swap Blocks} and \emph{Put Back Block}. For transfer, we evaluate RMBench-derived knowledge on RoboDojo~\citep{chen2026robodojo} using
\emph{Cover Blocks} and \emph{Press by Number}, with ten episodes per task. 

\paragraph{Agent models and comparisons.}
The main comparison uses Qwen3.8-27B, GPT-5.5, and GPT-6. Within each model, \emph{w/o HPK} uses the harness without persistent knowledge, while \emph{w/ Full HPK} uses both levels of HPK. We include $\pi_{0.5}$~\citep{physicalintelligence2025pi05} and Harness VLA~\citep{zhang2026harness} as reference methods.
The hierarchy ablations share source trajectories, the acting model, perception, executor, and action budget. \emph{Flat Reflection} summarizes the same source experience without separating task-level and action-level guidance. \emph{Task-only} and \emph{Action-only} use subsets of the same knowledge store. GPT-5.5 and GPT-6 are used for the correction, self-evolution, and transfer comparisons.

\paragraph{Evaluation protocol.}
We report task success and within-model gains in percentage points (pp). RoboDojo additionally reports a normalized task-progress Score. Evaluation uses fixed held-out configurations and read-only knowledge stores. Across-episode updates use the designated interaction trajectories, with VLM parameters and execution modules fixed. Appendix~\ref{sec:app_exp_protocol}
provides construction budgets, model configurations, metrics, and replication details.

\subsection{Real-World Experimental Setup}
\label{sec:app_real_world_setting}

\begin{figure}[!htbp]
    \centering
    \includegraphics[width=0.88\linewidth]{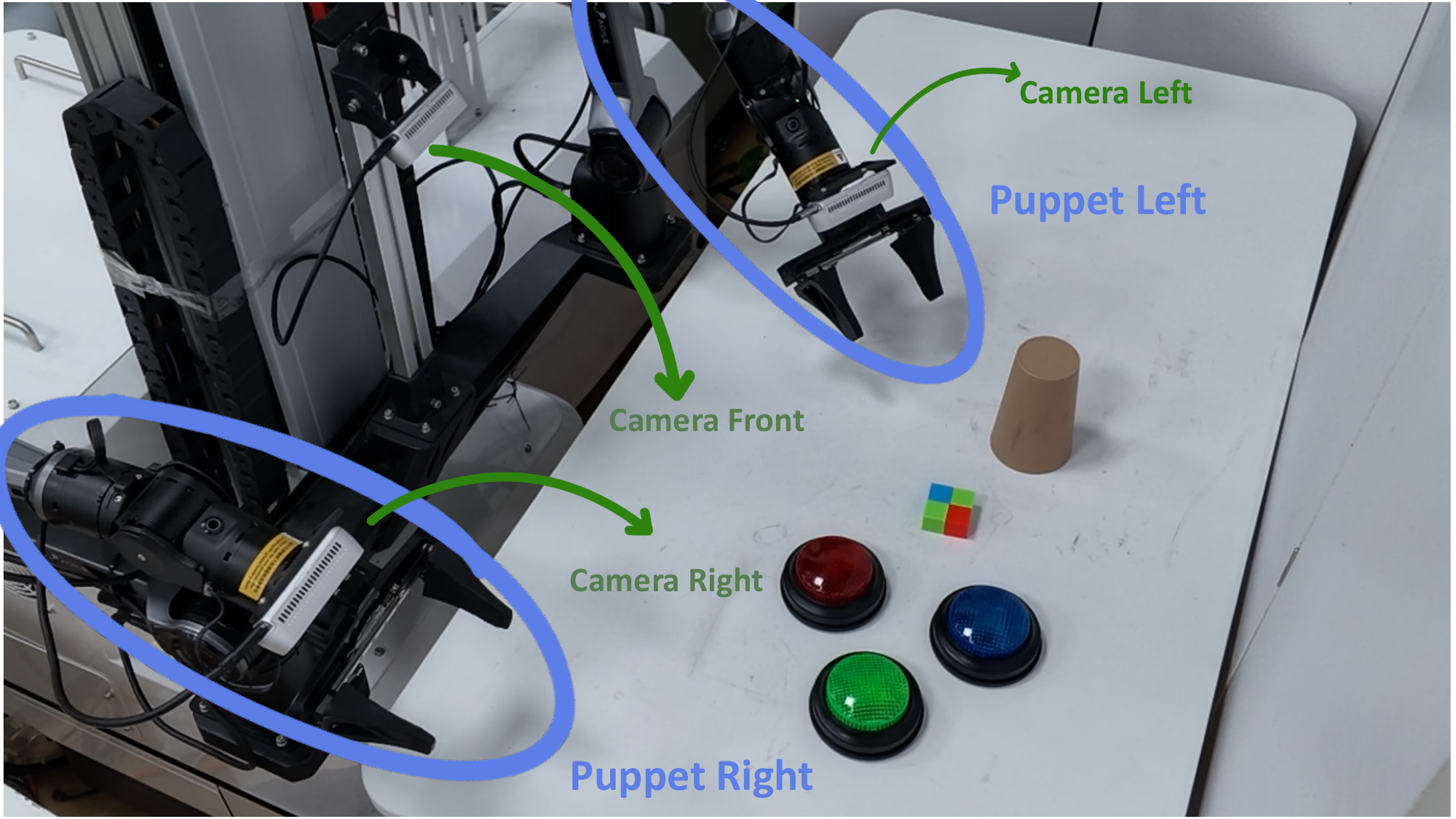}
    \caption{\textbf{Real-world hardware setup.}
    The AgileX Cobot Magic platform uses two follower arms for tabletop manipulation, with one front-view and two wrist-mounted RGB-D cameras. The follower arms are labeled \emph{Puppet Left} and \emph{Puppet Right} in the image.}
    \label{fig:app-agilex-cobot-magic}
\end{figure}

\paragraph{Hardware Setup.}
We conduct real-world experiments on an AgileX Cobot Magic platform configured as an ALOHA-style bimanual system~\citep{fu2024mobile,zhaolearning}. The platform comprises four 6-DoF Piper arms: two leader arms for human teleoperation and two follower arms for data collection and autonomous execution. Each follower arm is equipped with a parallel-jaw gripper. Three RealSense D435 RGB-D cameras provide a front view of the workspace and a wrist-mounted view from each follower arm. Fig.~\ref{fig:app-agilex-cobot-magic} shows the two follower arms and the camera locations; the leader arms are not shown.

\paragraph{$\pi_{0.5}$ baseline training and deployment.}
For the real-robot baseline, we initialize $\pi_{0.5}$ from the
pretrained \texttt{pi05\_base} weights and fine-tune all parameters
on 450 RMBench simulation demonstrations spanning nine tasks
(277,350 frames). The policy takes a task instruction, three RGB
views (workspace and two wrists), and a 14-dimensional bimanual
state comprising joint and gripper positions. It predicts
50-step action chunks. During training, joint targets are
represented relative to the current state, while gripper targets
remain absolute; output transforms recover absolute action targets.

Training uses eight GPUs with seed 42. We run 20,000 updates at
global batch size 64, then resume for 5,834 updates at batch size
256, totaling 25,834 updates. AdamW uses
$(\beta_1,\beta_2)=(0.9,0.95)$, $\epsilon=10^{-8}$, weight decay
$10^{-10}$, and gradient clipping at norm 1.0, with an exponential
moving average decay of 0.99. The learning-rate schedule uses
1,000 warmup steps to $2.5\times10^{-5}$, followed by cosine
decay toward $2.5\times10^{-6}$ over a configured total horizon
of 30,000 steps. We directly deploy the final checkpoint
(\texttt{25833}, indexed from zero) on the real robot without
any real-world fine-tuning. Policy parameters remain fixed during
evaluation.

\paragraph{Evaluation Rule.}
We evaluate three real-world tasks: \emph{Cover with Assigned Cups},
\emph{Press by Digit--Color Mapping}, and \emph{Uncover, Count, and Press}.
The comparison includes $\pi_{0.5}$, RoboHarn-Evo \emph{(Zero-shot)},
and RoboHarn-Evo \emph{(After RSI)}. Zero-shot uses HPK transferred
from simulation without prior real-world knowledge updates. After RSI
uses HPK further updated through real-world interaction, with the VLM
parameters and low-level executor fixed. Each RoboHarn-Evo setting is
evaluated on eight scenes per task. We report both full-task success
and normalized task progress to distinguish complete execution from
partial completion of these sequential tasks.

\paragraph{Task progress and success.}
Each task is divided into $M$ ordered subtasks, where $M=3$ for
\emph{Cover with Assigned Cups} and \emph{Press by Digit--Color Mapping},
and $M=4$ for \emph{Uncover, Count, and Press}.
Each subtask receives a score of $0$, $0.5$, or $1$:
\begin{itemize}
  \setlength{\itemsep}{0.15em}
  \setlength{\parsep}{0pt}
  \item $0$: No valid operation toward the required subtask has begun.
  \item $0.5$: The correct operation is underway, but the required
  outcome has not yet been completed and confirmed.
  \item $1$: The required outcome has been completed and confirmed.
\end{itemize}
These scores measure achieved task progress; the number of attempts
does not by itself reduce a completed subtask to $0.5$.
Scoring follows the required task order. A subtask receives no credit
unless all preceding subtasks have received $1$.
Using the wrong object or button, or failing to match the required
press count, does not qualify as completion.
Scores are reassessed if a previously completed state is disrupted;
the reported score reflects the state at the end of the rollout,
not the highest progress reached earlier.
Full task credit additionally requires compliance with all
task-specific ordering and arm-use constraints described below.

For rollout $n$, let $q_{n,j}$ denote the final score of subtask $j$.
The normalized rollout progress $P_n$ and the reported mean task
progress are
\begin{equation}
  P_n = \frac{1}{M}\sum_{j=1}^{M} q_{n,j},
  \qquad
  \mathrm{TaskProgress}
  = \frac{100}{N}\sum_{n=1}^{N} P_n,
  \label{eq:real_task_progress}
\end{equation}
where $N$ is the number of evaluated rollouts for the task.
Success requires the full task score, with all instruction constraints
satisfied. The success rate is the percentage of successful rollouts.
When aggregating across tasks, we average the three task-level
percentages with equal weight.

\paragraph{Task requirements and scoring criteria.}
The following criteria apply together with the shared ordering rule.
\begin{enumerate}
  \setlength{\itemsep}{0.7em}
  \setlength{\parsep}{0pt}

  \item \textbf{Cover with Assigned Cups.}
  The robot must cover the red, green, and blue blocks in that order.
  Each rollout instruction specifies which cup to use for each block.
  Cups are identified by their initial left, middle, or right position;
  these identities remain fixed after the cups move.
  The cup assignment is instruction-dependent, rather than a fixed
  left-to-right sequence. Both arms must be used during the task,
  and only one arm may move at a time.
  The maximum score is $3$.
  \begin{itemize}
    \setlength{\itemsep}{0.35em}
    \setlength{\parsep}{0pt}
    \item[-] \begin{samepage}\textit{Step 1: Covering the red block with the first designated cup.}
    \begin{itemize}
      \setlength{\itemsep}{0pt}
      \setlength{\parsep}{0pt}
      \item $0$: The robot has not begun valid manipulation of the designated cup.
      \item $0.5$: The designated cup has been grasped and is being
      transported or placed, but coverage of the red block is not confirmed.
      \item $1$: The red block is covered by the designated cup,
      and the cup has been released.
    \end{itemize}
    \end{samepage}
    \item[-] \begin{samepage}\textit{Step 2: Covering the green block with the second designated cup.}
    \begin{itemize}
      \setlength{\itemsep}{0pt}
      \setlength{\parsep}{0pt}
      \item $0$: The robot has not begun valid manipulation of the designated cup.
      \item $0.5$: The designated cup has been grasped and is being
      transported or placed, but coverage of the green block is not confirmed.
      \item $1$: The green block is covered by the designated cup,
      and the cup has been released.
    \end{itemize}
    \end{samepage}
    \item[-] \begin{samepage}\textit{Step 3: Covering the blue block with the third designated cup.}
    \begin{itemize}
      \setlength{\itemsep}{0pt}
      \setlength{\parsep}{0pt}
      \item $0$: The robot has not begun valid manipulation of the designated cup.
      \item $0.5$: The designated cup has been grasped and is being
      transported or placed, but coverage of the blue block is not confirmed.
      \item $1$: The blue block is covered by the designated cup,
      and the cup has been released.
    \end{itemize}
    \end{samepage}
  \end{itemize}
  Full completion requires all three assigned coverings to hold,
  with both arms used sequentially as instructed.

  \item \textbf{Press by Digit--Color Mapping.}
  The robot reads the left and right displayed digits.
  The instruction assigns each digit to a different button color.
  The robot first presses the button assigned to the left digit that
  many times, then presses the button assigned to the right digit
  that many times. It finally presses the remaining colored button
  exactly once to finish. The digit-to-color mapping is specified
  by the current instruction; the finishing button has no fixed color.
  The maximum score is $3$.
  \begin{itemize}
    \setlength{\itemsep}{0.35em}
    \setlength{\parsep}{0pt}
    \item[-] \begin{samepage}\textit{Step 1: Completing the presses specified by the left digit.}
    \begin{itemize}
      \setlength{\itemsep}{0pt}
      \setlength{\parsep}{0pt}
      \item $0$: No valid pressing of the assigned button has begun.
      \item $0.5$: The robot is pressing the correct button, but the
      required count has not yet been completed and confirmed.
      \item $1$: The assigned button has been pressed exactly as many
      times as specified by the left digit.
    \end{itemize}
    \end{samepage}
    \item[-] \begin{samepage}\textit{Step 2: Completing the presses specified by the right digit.}
    \begin{itemize}
      \setlength{\itemsep}{0pt}
      \setlength{\parsep}{0pt}
      \item $0$: No valid pressing of the assigned button has begun.
      \item $0.5$: The robot is pressing the correct button, but the
      required count has not yet been completed and confirmed.
      \item $1$: The assigned button has been pressed exactly as many
      times as specified by the right digit.
    \end{itemize}
    \end{samepage}
    \item[-] \begin{samepage}\textit{Step 3: Pressing the remaining button to finish.}
    \begin{itemize}
      \setlength{\itemsep}{0pt}
      \setlength{\parsep}{0pt}
      \item $0$: No valid pressing of the remaining button has begun.
      \item $0.5$: The robot is pressing the remaining button, but
      completion of the press is not yet confirmed.
      \item $1$: The remaining button has been pressed exactly once
      after both digit-specified sequences are complete.
    \end{itemize}
    \end{samepage}
  \end{itemize}

  \item \textbf{Uncover, Count, and Press.}
  The robot first uses one arm to remove the cup covering the blocks
  and release it in a safe location. It then observes the numbers of
  red, green, and blue blocks and uses the other arm to press the
  corresponding buttons in red--green--blue order.
  Each button must be pressed exactly as many times as there are
  blocks of that color. Cup removal and button pressing must be
  performed sequentially by different arms.
  No additional finishing-button press is required.
  Counting determines the required press counts and is not scored
  as a separate subtask. The maximum score is $4$.
  \begin{itemize}
    \setlength{\itemsep}{0.35em}
    \setlength{\parsep}{0pt}
    \item[-] \begin{samepage}\textit{Step 1: Removing and safely placing the cup.}
    \begin{itemize}
      \setlength{\itemsep}{0pt}
      \setlength{\parsep}{0pt}
      \item $0$: The cup still covers the blocks, and valid removal
      has not begun.
      \item $0.5$: The cup has been grasped and is being lifted,
      transported, or placed, but safe placement is incomplete.
      \item $1$: The cup has been released in a safe location,
      leaving all blocks visible.
    \end{itemize}
    \end{samepage}
    \item[-] \begin{samepage}\textit{Step 2: Pressing the red button to match the red-block count.}
    \begin{itemize}
      \setlength{\itemsep}{0pt}
      \setlength{\parsep}{0pt}
      \item $0$: No valid pressing of the red button has begun.
      \item $0.5$: The robot is pressing the red button, but the
      required count has not yet been completed and confirmed.
      \item $1$: The red button has been pressed exactly as many
      times as there are red blocks.
    \end{itemize}
    \end{samepage}
    \item[-] \begin{samepage}\textit{Step 3: Pressing the green button to match the green-block count.}
    \begin{itemize}
      \setlength{\itemsep}{0pt}
      \setlength{\parsep}{0pt}
      \item $0$: No valid pressing of the green button has begun.
      \item $0.5$: The robot is pressing the green button, but the
      required count has not yet been completed and confirmed.
      \item $1$: The green button has been pressed exactly as many
      times as there are green blocks.
    \end{itemize}
    \end{samepage}
    \item[-] \begin{samepage}\textit{Step 4: Pressing the blue button to match the blue-block count.}
    \begin{itemize}
      \setlength{\itemsep}{0pt}
      \setlength{\parsep}{0pt}
      \item $0$: No valid pressing of the blue button has begun.
      \item $0.5$: The robot is pressing the blue button, but the
      required count has not yet been completed and confirmed.
      \item $1$: The blue button has been pressed exactly as many
      times as there are blue blocks.
    \end{itemize}
    \end{samepage}
  \end{itemize}
  Full completion requires safe cup placement and all three press
  sequences, with the instructed separation of arm roles.
\end{enumerate}

\begin{figure*}[t]
\centering
\includegraphics[width=\textwidth]{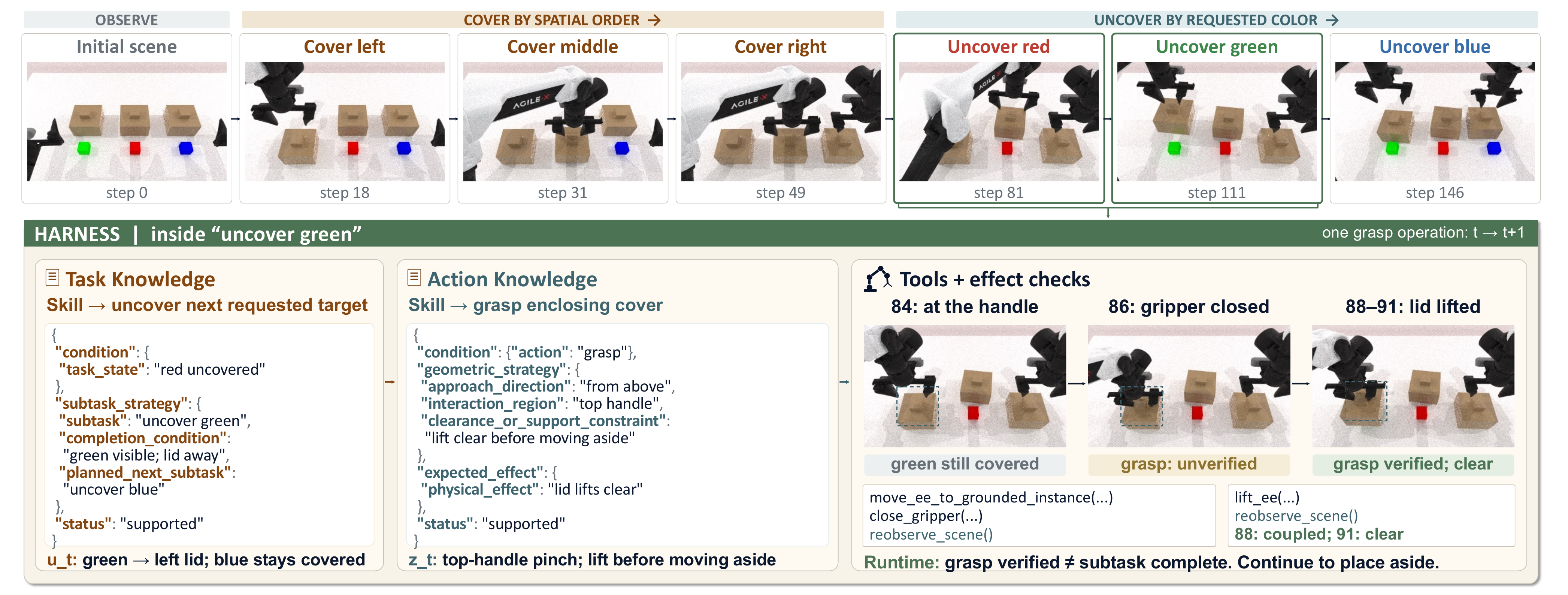}
\caption{Example of Cover Blocks.}
\label{fig:episode_walkthrough}
\end{figure*}

\graphicspath{{Figure/}}
\definecolor{AFKTaskBg}{HTML}{F2F7FF}
\definecolor{AFKActionBg}{HTML}{F4FAF4}
\newcolumntype{Y}{>{\raggedright\arraybackslash}X}

\section{Examples of Persistent Task and Action Knowledge}
\label{app:knowledge-examples}

We provide representative entries from the persistent knowledge store for the
\emph{Cover Blocks} task. The source package was extracted from ten fixed
demonstration trajectories and contains six Task Knowledge entries grouped into
two Task Skills, and four Action Knowledge entries grouped into four Action
Skills. The examples below preserve the semantic fields stored by the method;
identifiers and runtime bindings are intentionally omitted because they are not
part of the VLM-facing knowledge representation.
Table~\ref{tab:cover-blocks-knowledge-catalog}
summarizes the complete persistent store for this task.
Tables~\ref{tab:task-knowledge-example} and~\ref{tab:action-knowledge-example}
then show one Task Knowledge entry and its complementary Action Knowledge entry,
respectively, while Figure~\ref{fig:knowledge-example-evidence} grounds both
entries in the corresponding execution sequence.

\begin{table*}[t]
  \centering
  \caption{Complete catalog of the persistent knowledge entries extracted for
  \emph{Cover Blocks}. The table summarizes semantic content and skill grouping,
  rather than treating the independently collected HPK examples as repeated
  validation trials.}
  \label{tab:cover-blocks-knowledge-catalog}
  \small
  \setlength{\tabcolsep}{4pt}
  \begin{tabularx}{\textwidth}{@{}llY@{}}
    \toprule
    Type & Skill family & Persistent entry \\
    \midrule
    Task & Cover by spatial order & Cover the leftmost visible target \\
    Task & Cover by spatial order & Cover the next (middle) visible target \\
    Task & Cover by spatial order & Cover the remaining rightmost target \\
    Task & Uncover by requested identity & Uncover the red block first \\
    Task & Uncover by requested identity & Uncover the green block next \\
    Task & Uncover by requested identity & Uncover the blue block last \\
    \midrule
    Action & Grasp free cover & Grasp a table-supported lid by its exposed handle \\
    Action & Grasp enclosing cover & Grasp a covering lid and lift it clear of the block \\
    Action & Place cover to enclose & Center a held lid over a visible block and release \\
    Action & Place removed cover aside & Set a removed lid down away from the revealed block \\
    \bottomrule
  \end{tabularx}
\end{table*}

\subsection{Task Knowledge example}

Table~\ref{tab:task-knowledge-example} shows how Task Knowledge determines
\emph{what to do next} from task progress and the requested semantic order. It
deliberately selects the target by identity rather than by spatial position. In
Figure~\ref{fig:knowledge-example-evidence}(a,c), all blocks are initially
covered, and the red block is visible after the selected lid has been moved away.

\begin{table*}[t]
  \centering
  \caption{Representative Task Knowledge entry (semantic content preserved,
  reformatted for readability).}
  \label{tab:task-knowledge-example}
  \small
  \colorbox{AFKTaskBg}{%
  \parbox{\dimexpr\textwidth-2\fboxsep\relax}{%
  \vspace{2pt}
  \begin{tabularx}{\linewidth}{@{}>{\ttfamily\bfseries}p{0.28\linewidth}Y@{}}
    overall\_goal & Cover all targets from left to right, then uncover them in red, green, blue order. \\
    task\_state & Start of the uncovering phase after all targets have been covered. \\
    relevant\_relations & All target blocks are covered; the red block is under one lid; the lid over the red block is reachable. \\
    \addlinespace[2pt]
    subtask & Uncover the red block. \\
    purpose & Begin the requested uncovering sequence with red. \\
    selection\_basis & Select the lid covering the red block, independent of spatial position, because red is first in the requested uncover order. \\
    completion\_condition & The red block is visible and its lid has been moved away while later-color blocks remain covered. \\
    planned\_next\_subtask & Uncover the green block. \\
  \end{tabularx}
  \vspace{2pt}}}
\end{table*}

\subsection{Action Knowledge example}

\begin{figure*}[t]
  \centering
  \begin{minipage}[t]{0.31\textwidth}
    \centering
    \includegraphics[width=\linewidth]{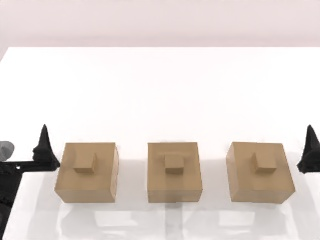}\\[-1pt]
    \footnotesize (a) Before: all blocks are covered.
  \end{minipage}\hfill
  \begin{minipage}[t]{0.31\textwidth}
    \centering
    \includegraphics[width=\linewidth]{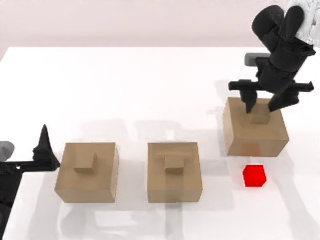}\\[-1pt]
    \footnotesize (b) Action effect: the red block becomes visible while the lid is held.
  \end{minipage}\hfill
  \begin{minipage}[t]{0.31\textwidth}
    \centering
    \includegraphics[width=\linewidth]{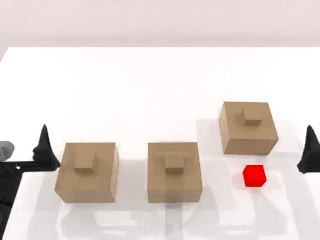}\\[-1pt]
    \footnotesize (c) Subtask complete: the lid is set aside and the red block remains visible.
  \end{minipage}
  \caption{Evidence associated with the two representative entries from
  demonstration \texttt{episode\_9}. The Task Knowledge evidence spans frames
  512--687. The Action Knowledge evidence for grasping the enclosing lid spans
  frames 512--600. The final frame also verifies completion of the subsequent
  set-aside action.}
  \label{fig:knowledge-example-evidence}
\end{figure*}

Table~\ref{tab:action-knowledge-example} specifies \emph{how to realize} the
selected subtask. Its applicability condition distinguishes removing an
enclosing lid from grasping a free-standing lid, while its expected effect
provides an observable verification criterion. The intermediate observation in
Figure~\ref{fig:knowledge-example-evidence}(b) directly verifies this expected
effect: the red block becomes visible while the lid moves with the gripper.

\begin{table*}[t]
  \centering
  \caption{Representative Action Knowledge entry (semantic content preserved,
  reformatted for readability).}
  \label{tab:action-knowledge-example}
  \small
  \colorbox{AFKActionBg}{%
  \parbox{\dimexpr\textwidth-2\fboxsep\relax}{%
  \vspace{2pt}
  \begin{tabularx}{\linewidth}{@{}>{\ttfamily\bfseries}p{0.28\linewidth}Y@{}}
    action & grasp. \\
    object\_description & A lid currently covering a block. \\
    held\_state & No lid held. \\
    support\_relation & The lid is supported on the table around or over the block. \\
    target\_relation & The top handle is exposed and reachable. \\
    \addlinespace[2pt]
    approach & From above, referenced to the covering lid's top handle. \\
    interaction & Close the fingers around opposite sides of the handle; align the gripper span across the handle and keep the lid level. \\
    clearance & Lift vertically high enough to clear the hidden block before moving the lid laterally away. \\
    avoid & The block beneath the lid; neighboring covered or set-aside lids; nearby exposed blocks; workspace edges when near an edge. \\
    \addlinespace[2pt]
    physical\_effect & The gripper gains control of the covering lid and lifts it clear of the block. \\
    verification & The covered block becomes visible while the lid rises with the gripper. \\
  \end{tabularx}
  \vspace{2pt}}}
\end{table*}

\paragraph{Example provenance and scope.}
The frames in Figure~\ref{fig:knowledge-example-evidence} provide traceability
for this independently collected HPK example: they show the state transition
from the applicable condition to the expected effect and subtask completion.
They are illustrative observations from the same execution sequence, not an
aggregate reliability evaluation or repeated validation trial. This example
demonstrates the extracted knowledge structure and its grounding in a fixed
demonstration; it does not by itself establish real-robot effectiveness.

\section{Additional Method Details}
\label{sec:app_method}

RoboHarn-Evo evolves the knowledge that guides a fixed VLM and robot
executor. Its central mechanism is a shared subtask context: Task Knowledge
specifies the objective, Action Knowledge specifies how to act toward it,
and physical evidence determines how both are revised before reuse.
This appendix details the representation, retrieval, and maintenance
procedures that implement this interaction--update cycle.

\subsection{Hierarchical knowledge representation}
\label{sec:app_method_knowledge}
\label{sec:app_method_notation}

\paragraph{Runtime context and persistent knowledge.}
Within episode $i$, the harness maintains $s_t$ from observations and
interaction history. It records the active subtask, relevant objects and
relations, executed operations, and their observed outcomes. Across episodes,
$\mathcal K_i=(\mathcal K_i^{\mathrm T},\mathcal K_i^{\mathrm A})$ stores
reusable Task and Action Knowledge, and
$\mathcal F_i=(\mathcal F_i^{\mathrm T},\mathcal F_i^{\mathrm A})$ indexes
these entries by Skill. Following Sec.~\ref{sec:problem_formulation}, $t$
indexes operation boundaries, $c_t$ denotes parameterized tool calls, and
$a_t$ denotes the resulting robot-action sequence. A subtask can remain
active across several operations; its completion is assessed separately
from the success of each operation.

\paragraph{Task and Action entries.}
A Task entry describes the conditions under which a subtask is appropriate,
its purpose, its selection rationale, and its completion condition. An
Action entry describes an operation type, applicable object and support
relations, an object-relative geometric strategy, and its intended physical
effect. Geometric strategies specify approach direction, contact region,
gripper orientation, and clearance or support requirements. They are
instantiated using current observations rather than stored robot poses.
Both entry types retain links to their source executions, an evidence
summary $\mathbf b_k=(n_k^+,n_k^-,n_k^?)$, and a retrieval status
$\sigma(k)$. Table~\ref{tab:hpk_example} illustrates the two levels using
\emph{Cover Blocks}.

\begin{table}[!htbp]
\centering
\caption{Two linked knowledge entries for \emph{Cover Blocks}, with the
requested uncovering order red, green, then blue. The grasp effect is a
step toward, rather than the completion of, the uncovering subtask.}
\label{tab:hpk_example}
\small
\setlength{\tabcolsep}{4pt}
\renewcommand{\arraystretch}{1.12}
\begin{tabularx}{\linewidth}{@{}p{0.14\linewidth}XX@{}}
\toprule
Component & Task Knowledge & Action Knowledge \\
\midrule
Condition & Red is uncovered; green and blue remain covered.
& A lid covers the target block; its top handle is reachable. \\
Strategy & Uncover green next, according to the requested color order.
& Pinch opposite sides of the top handle; lift vertically to clear the block
before moving laterally. \\
Outcome & Green is visible and its lid is moved away; blue remains covered.
& The lid rises with the gripper and clears the block. \\
Evidence & Observations at the subtask boundaries establish the required
change in task state.
& Observations of the grasp and lift establish control of the lid. \\
\bottomrule
\end{tabularx}
\end{table}

\paragraph{From an episode to a knowledge package.}
The reflector receives $\operatorname{View}(\tau_i)$: observations, tool
calls, and execution feedback associated with the episode. It constructs
$\mathcal P_i$ as in Eq.~\eqref{eq:afk-package}, preserving the order of
subtasks $k_{i,j}^{\mathrm T}$ and the operations
$\{k_{i,j,r}^{\mathrm A}\}_{r=1}^{m_{i,j}}$ within each subtask. Before-and-after
observations remain associated with the operations they assess. Observed
facts and inferred rationales are recorded separately, so a proposed
explanation does not replace the physical outcome of an attempt. Atomic
entries are the units of retrieval and revision; the package retains their
shared task context and temporal associations.

\paragraph{Skill organization.}
A Skill groups entries that address a related decision context and
summarizes their applicability. For example, color-order uncovering groups
Task entries, while lid grasping and lid placement group Action entries.
A group can contain distinct strategies or conditions. Group membership
therefore determines where retrieval and maintenance look; semantic
consolidation determines which entries express the same knowledge.

\subsection{Hierarchical retrieval and goal-consistent execution}
\label{sec:app_method_retrieval}

\paragraph{Selecting what to accomplish.}
Task retrieval uses the instruction, current runtime memory, and the
planner's baseline subtask. Skill summaries identify relevant groups, from
which the VLM selects supported entries whose conditions match the current
state. These entries refine $u_t$, including the designated object, target,
and completion condition. If no entry applies, the planner retains its
baseline decision. In the example of Table~\ref{tab:hpk_example}, the
requested color sequence determines the next target after red is uncovered,
regardless of the targets' left-to-right arrangement.

\paragraph{Selecting how to act.}
Action retrieval follows the selection of $u_t$ and operation type
$\alpha_t\in\{\texttt{grasp},\texttt{place},\texttt{contact}\}$. The query
includes the target object, held state, support relations, and descriptions
of the current candidate geometry. The VLM selects applicable Action
Knowledge together with a geometric realization $z_t$. Its conditions must
match the present interaction: grasping a lid that covers a block, for
example, requires clearance above the hidden block before lateral motion.
The retrieved strategy supplies geometric guidance within the selected
subtask rather than changing its target.

\paragraph{Skill routing.}
The routing defaults use exhaustive recall for stores of at most 24 entries.
For larger stores, local inverse-document-frequency overlap ranks Skill
summaries and member cards; up to two Skills are selected, and their members
are interleaved to form a shortlist of up to eight entries. When lexical
overlap is uninformative, the VLM selects Skills from semantic routing
cards. On this model-routed path, up to 24 members can be passed directly
to the final selector; larger candidate sets are first shortlisted to eight.
Both levels filter for supported entries, and Action retrieval additionally
matches the operation type. Final selection uses compact semantic cards
and the current scene context.

\paragraph{Grounding under the task goal.}
\label{sec:app_method_goal_implementation}
The harness generates candidate poses from current observations and checks
feasibility together with the object, target, and target relation specified
by $u_t$. Candidates matching $z_t$ are prioritized within this admissible
set, preserving their relative order. The VLM issues tool calls $c_t$, and
scene grounding, motion planning, and control produce $a_t$. Without
applicable Action Knowledge, the baseline candidate ranking is retained
under the same task constraints. If no candidate satisfies these constraints,
the agent observes again or replans instead of substituting a different
target.

\paragraph{Closing the execution loop.}
Execution checks and VLM assessment compare the observed outcome with the
intended effect. Their result updates $s_t$ and informs whether to continue,
re-observe, or change the attempted realization. The planner advances to
the next subtask when its completion condition is verified. Thus control of
a grasped object can enable a subsequent transport operation while leaving
the placement or uncovering subtask active. Tool-call completion, operation
effect, and subtask completion remain distinct throughout the trajectory.

\subsection{Physical evidence and knowledge revision}
\label{sec:app_method_reflection_implementation}
\label{sec:app_method_maintenance}

\paragraph{Evidence is tied to the claim being tested.}
Each judgment concerns a particular knowledge entry, its content version,
and a source execution. Assessment first establishes that the entry's
condition held and that the executed operation followed its strategy under
the designated task goal. The outcome is then labeled \texttt{support} when
the intended effect is observed, \texttt{oppose} when the observations
establish failure, and \texttt{unverified} when execution deviated or the
available observations are inconclusive. An out-of-scope execution does
not contribute evidence to that entry. This attribution keeps an off-target
local success from reinforcing a strategy for the intended task.

\paragraph{Verification follows the knowledge hierarchy.}
Action evidence concerns the physical effect of an operation, such as an
object moving with the gripper after a grasp, resting on its target support
after release, or changing state after contact. Task evidence concerns the
completion condition of the enclosing subtask and the required task
relations. It is assessed at subtask boundaries rather than inferred by
summing successful constituent operations. A successful grasp can therefore
support Action Knowledge while the associated Task entry awaits evidence
that its completion condition has been reached.

\paragraph{Delayed observations refine the same attempt.}
Consider removing the lid over the green block in
Table~\ref{tab:hpk_example}. Gripper closure alone can leave attachment
uncertain. A subsequent lift showing the lid moving with the gripper
supports the grasp strategy. The uncovering subtask is completed when green
is visible and the lid has been moved away, with blue still covered.
These observations address different levels of the hierarchy. At the
operation level, the lift refines the pending grasp judgment instead of
creating a second grasp trial. Source associations preserve one verdict
per attempt for each assessed content version. If a pending verdict has
already been recorded, resolving it replaces that verdict rather than
appending a duplicate event.

\paragraph{Evidence summaries and retrieval eligibility.}
For an entry $k$, $n_k^+$, $n_k^-$, and $n_k^?$ count distinct applicable
attempts labeled support, opposition, and unverified, respectively. The
retrieval status is determined by
\begin{equation}
\sigma(k)=
\begin{cases}
\texttt{contested}, & n_k^->0,\\
\texttt{supported}, & n_k^+>0\ \text{and}\ n_k^-=0,\\
\texttt{candidate}, & n_k^+=n_k^-=0.
\end{cases}
\label{eq:afk-status}
\end{equation}
Retrieval considers supported entries whose conditions apply to the current
decision. Unverified outcomes remain in the evidence record without
increasing support or opposition. A newly proposed strategy has candidate
status until an applicable execution supplies support. Counter-evidence
withdraws the challenged version from retrieval and directs maintenance
toward its content and applicability conditions.

\paragraph{Consolidating equivalent knowledge.}
New atoms are assigned to existing Skills, or to a new Skill when their
decision context is not represented. Maintenance then operates on the
corresponding Task or Action partitions, including historical entries in
the affected Skills. Entries are consolidated when their conditions,
strategies, and intended effects are equivalent. Entries that share an
object or operation but prescribe different interactions remain distinct.
Evidence summaries are combined from the associated attempts, with repeated
references to the same attempt counted once. Runtime and reflected records
of an operation therefore contribute the same evidence event. Consolidation
preserves supporting, opposing, and unresolved observations together.

\paragraph{Revising challenged content.}
Maintenance uses before-and-after observations and execution records to
review historical entries alongside new knowledge. A revision changes the
applicable condition or recommended strategy in response to the observed
outcomes. Narrowing an entry to a demonstrated interaction context and
changing its grasp geometry are different revisions: each produces a
content version whose claims are checked against relevant executions.
Evidence is associated with the version it assesses. For a narrower
condition, an earlier execution contributes only when that condition held;
for a changed strategy, it contributes only when the changed strategy was
actually executed. A rewritten description therefore receives support
through an execution that tests it, not through the act of rewriting.

The revised version becomes retrievable when it satisfies
Eq.~\eqref{eq:afk-status}. Earlier versions retain their evidence, including
opposition. This separates withdrawal from repair: withdrawal prevents
reuse of a challenged claim, whereas repair establishes an evidence-supported
replacement with revised content. A promising but untested revision remains
a candidate. An observed failure that still applies to the revised claim
remains opposing evidence for that claim.

\paragraph{Refreshing the persistent store.}
After consolidation and revision, the maintainer refreshes the summaries
and membership of the affected Skills. Unaffected entries remain unchanged.
The updated $(\mathcal K_{i+1},\mathcal F_{i+1})$ supplies the next episode's
retrieval, so later decisions reflect both newly supported strategies and
changes to historical knowledge. Physical feedback can thus change what
the harness recommends and the conditions under which it recommends it,
while the base VLM and execution tools remain fixed.

\begin{samepage}
\subsection{The interaction--update cycle}
\label{sec:app_method_evolution}

Algorithm~\ref{alg:hpk_cycle} combines the procedures above. During an
episode, observations and effect checks update runtime memory and provide
the evidence for the next knowledge update. After the episode, reflection
and maintenance update the persistent store. The next trajectory is
therefore generated using knowledge shaped by earlier trajectories,
closing the recursive loop between execution and knowledge revision.
\par
\end{samepage}

\begin{algorithm}[!htbp]
\caption{RoboHarn-Evo: execution and evidence-driven knowledge evolution}
\label{alg:hpk_cycle}
\begin{algorithmic}[1]
\Require Fixed VLM parameters $\theta$, executor $\operatorname{Exec}$,
initial store $\mathcal K_0$ and Skill index $\mathcal F_0$
\For{each interaction episode $i$ with instruction $l$}
    \State Initialize $s_0$ from the initial observations; set $t=0$.
    \While{the task is active and the execution budget remains}
        \State Select or retain $u_t$ using Task retrieval and verified task progress.
        \State Choose $\alpha_t$; retrieve Action Knowledge to select $z_t$ for $u_t$.
        \State Ground $z_t$ using feasible, goal-consistent scene candidates.
        \If{no admissible realization is available}
            \State Re-observe or replan; record the unresolved attempt.
        \Else
            \State Issue tool calls $c_t$ and execute robot actions $a_t$.
            \State Assess the operation effect and subtask completion separately.
            \State Retain observations, tool calls, and effect judgments.
        \EndIf
        \State Update runtime memory $s_{t+1}$ from feedback; increment $t$.
    \EndWhile
    \State Reflect on $\operatorname{View}(\tau_i)$ to construct $\mathcal P_i$.
    \State Assess new Task and Action entries against level-specific physical evidence.
    \State Assign new atoms to Skills and collect affected historical entries.
    \State Review historical entries; consolidate equivalents and revise challenged content.
    \State Check revisions against relevant executions; update version-specific evidence.
    \State Apply Eq.~\eqref{eq:afk-status} and refresh affected Skill summaries.
    \State Retain $(\mathcal K_{i+1},\mathcal F_{i+1})$ for subsequent episodes.
\EndFor
\end{algorithmic}
\end{algorithm}

\section{Detailed experimental protocols and result tables}
\label{sec:app_exp_protocol}

\FloatBarrier
\subsection{Q2: experience-pool self-correction}
\label{sec:app_q2_audit}

\subsubsection{Metric definitions}
\label{sec:app_q2_metrics}

Let $\mathcal V_K$ be the active pool and $\mathcal R_K\subseteq\mathcal V_K$
the reviewed entries at checkpoint $K$. Let $C_K$, $I_K$, and $U_K$ count
correct, incorrect, and unverified labels in $\mathcal R_K$, with
$N_K=C_K+I_K+U_K$. Then
\begin{equation}
\mathrm{CER}_K=100\frac{C_K}{N_K},\qquad
\mathrm{UR}_K=100\frac{U_K}{N_K}.
\label{eq:app_cer}
\end{equation}
For a census, $\mathcal R_K=\mathcal V_K$; for a sampled review, the reported
ratio is over reviewed active entries, accompanied by the sampling design
and coverage. Unequal-probability sampling requires the corresponding weighted
pool estimate. Unverified entries remain in the denominator; unaudited entries
are neither silently labeled unverified nor counted as correct.

Let $\mathcal H_0^-$ and $\mathcal H_0^+$ be the fixed initially reviewed
incorrect and correct cohorts. Each initially incorrect claim has exactly one
current outcome: \emph{repaired}, \emph{inactive}, \emph{still incorrect}, or
\emph{unverified}. Repair requires the relevant active descendants to be
independently reviewed as correct; inactivation requires that no active
descendant retains the erroneous claim. A remaining erroneous branch prevents
a split or merge from counting as resolved. If $R_K^{\mathrm{fix}}$ and
$R_K^{\mathrm{off}}$ count the first two outcomes,
\begin{equation}
\mathrm{HERR}_K=100\frac{R_K^{\mathrm{fix}}+R_K^{\mathrm{off}}}
{|\mathcal H_0^-|}.
\label{eq:app_herr}
\end{equation}
The two terms are reported separately. For initially valid knowledge, let
$T_K$ count claims with an active, correct semantic successor retaining their
original valid scope. Correct retention is
\begin{equation}
\mathrm{CERet}_K=100\frac{T_K}{|\mathcal H_0^+|}.
\label{eq:app_retention}
\end{equation}

\FloatBarrier
\subsection{Q4: frozen-source transfer and target adaptation}
\label{sec:app_transfer}

RoboDojo evaluates both partial task progress and terminal success.
For episode $e$, let $r_e$ denote the task score. We report
\begin{equation}
\mathrm{Score}
=
\frac{100}{N}\sum_{e=1}^{N} r_e,
\qquad
\mathrm{SR}
=
\frac{100}{N}\sum_{e=1}^{N}
\mathbbm{1}[r_e=1],
\label{eq:robodojo_metrics}
\end{equation}
with $N=10$ episodes for each transfer condition.

For Cover Blocks,
\[
r_e\in\{0,\;0.05,\;0.15,\;0.30,\;1.0\}.
\]
The intermediate values correspond to the furthest verified stage reached:
all three blocks covered ($0.05$), red uncovered ($0.15$), and red and green
uncovered while blue remains covered ($0.30$). A score of $1.0$ requires
completion of the full sequence and return of both arms.

For Press by Number,
\[
r_e\in\{0,1\},
\]
because the task exposes terminal success only. Its mean Score therefore
equals its Success Rate.

\section{Runtime Skill Library}
\label{app:runtime_skills}

\subsection{Overview and Organization}
\label{app:runtime_skills_overview}

The agent is equipped with a library of \emph{runtime skills} that define structured behavioral contracts for different stages of the manipulation runtime. Each skill specifies the information available to the agent, the responsibility of the corresponding runtime step, the expected output format, and the constraints that must be respected. These skills are prompt-level runtime interfaces rather than additional learned policies or model parameters.

The library contains 28 skills organized into six functional categories: Perception, Memory, Planning, Monitoring, Tool Calling, and Effect Verification. Perception, Memory, Monitoring, and Effect Verification each contain a single dedicated skill. Planning contains five skills for different planning and runtime-reasoning responsibilities. Tool Calling contains 19 skills, organized into one routing skill, ten reusable primitive skills, and eight workflow skills. Experience records, including raw traces, case summaries, learned lessons, and retrieval artifacts, are maintained separately from the runtime skill prompts and are not counted as runtime skills.

\begin{table}[t]
\centering
\small
\begin{tabular}{lcp{0.58\linewidth}}
\toprule
\textbf{Category} & \textbf{\# Skills} & \textbf{Primary Responsibility} \\
\midrule
Perception & 1 & Normalize perception queries before segmentation and scene-memory binding. \\
Memory & 1 & Convert observation and execution evidence into compact runtime memory. \\
Planning & 5 & Maintain task planning, runtime reasoning, and high-level execution decisions. \\
Monitoring & 1 & Detect execution health and semantic OOD states. \\
Tool Calling & 19 & Route tool use and construct grounded tool-call procedures from reusable primitives and workflows. \\
Effect Verification & 1 & Verify whether executed actions produced their intended physical effects. \\
\midrule
\textbf{Total} & \textbf{28} & \\
\bottomrule
\end{tabular}
\caption{Organization of the runtime skill library.}
\label{tab:runtime_skill_overview}
\end{table}

The categories define complementary runtime responsibilities rather than independent policies. Perception normalizes task-conditioned queries used by downstream grounding. Memory summarizes task-relevant observation and execution evidence into existing runtime memory surfaces. Planning maintains task and subtask decisions from the available runtime state. Monitoring detects execution anomalies and produces explicit semantic OOD information. Tool Calling maps the current execution context to structured tool-use procedures through routing, workflows, and reusable primitives. Effect Verification evaluates the observed physical consequence of an executed action or tool sequence.

We distinguish these runtime skills from the \emph{semantic Skills} used by Action-Feedback Knowledge (AFK) in the main method. AFK Skills organize atomic Task or Action Knowledge for retrieval, whereas the runtime skills described here define the structured interfaces used by the agent during perception, memory construction, planning, monitoring, tool calling, and effect verification.

\subsection{Perception Skills}
\label{app:runtime_skills_perception}

The Perception category contains a single skill for normalizing perception requests before segmentation and scene-memory binding.

\begin{skillbox}{Perception Query Normalization.}

\textbf{Purpose and Invocation.} This skill converts raw perception-query objects into a stable semantic interface for downstream perception. It is applied before segmentation and scene-memory binding, and is responsible for normalizing object categories, task roles, relational query types, and optional instance bindings. When a previous query set is rejected by the runtime, the same skill can regenerate the queries using the supplied failure reason and binding requirements.

\textbf{Inputs.} The skill receives the current task context, committed memory, observation summary, raw perception queries, and the candidate identities currently advertised by the runtime. It also receives the current instance-binding phase, whether exact binding is required, an optional binding postcondition describing a rejected previous result, and the maximum number of queries that may be returned. In simulator environments, an optional oracle-object catalog may also be provided. Each raw query may specify an object or part phrase, a segmentation prompt, a task role, an entity scope, an optional spatial relation, instance-level disambiguation, and an optional advertised runtime or oracle identity.

\textbf{Core Procedure.}
\begin{enumerate}
    \item Normalize each query into a stable object or part category while preserving descriptors that are useful for segmentation.
    \item Determine whether the query serves as a target, tool, or contextual reference, and separate category-level semantics from instance-level disambiguation.
    \item Represent ordinary objects as single-instance queries. When multiple visual references jointly define a placement relation, preserve them as a reference set rather than converting the relation into a metric coordinate.
    \item Resolve instance identity according to the current binding phase. Candidate discovery permits a task-supported target or tool to remain unbound when no matching candidate has yet been detected, whereas execution-facing selection requires an exact advertised identity when binding is enabled.
    \item If a previous query set was rejected, use the reported failure reason and previous output to correct the missing role or identity rather than repeating the rejected result unchanged.
\end{enumerate}

\textbf{Output Contract.} The skill returns JSON only. The output contains at most the runtime-specified maximum number of normalized queries and follows the schema below.

\begin{Verbatim}[breaklines=true,breakanywhere=true]
{
  "queries": [
    {
      "object_id": "<category>",
      "text_prompt": "<segmentation prompt>",
      "role": "target | tool | context",
      "entity_scope": "single_instance | reference_set",
      "placement_relation": "center_of | omitted",
      "expected_count": "<int | omitted>",
      "instance_hint": "<descriptor | empty>",
      "instance_ref": "<track_id / instance_id | empty>",
      "oracle_id": "<oracle_id | empty>",
      "reason": "<short rationale>"
    }
  ]
}
\end{Verbatim}

\textbf{Key Constraints.}
\begin{itemize}
    \item Object categories should remain stable across frames, while instance-specific descriptors are represented separately.
    \item Objects and identities that are not supported by the task, raw queries, or advertised candidates must not be invented.
    \item Exact runtime or oracle identifiers must be preserved without shortening or rewriting, and candidates must not be selected by list order.
    \item A visual reference set remains a relational query. The skill does not choose an arbitrary member of the set or generate the corresponding metric placement coordinate.
    \item Candidates retained only for recovery-time identity binding cannot provide current geometric evidence for grounded task progress.
    \item If a required target or tool cannot be identified confidently, the skill omits the unsupported query rather than guessing. If the required role itself cannot be supported, an empty query list is allowed.
\end{itemize}

\textbf{Interaction with Other Components.} This skill receives raw queries from the perception-query planner and produces normalized queries for downstream segmentation and scene-memory binding. The runtime validates the returned schema, required roles, identifier membership, grounding validity, visibility or retained-track status, supported relations, and query count before the queries are used downstream. Relational reference sets are converted by the runtime into separately verified placement targets rather than metric poses produced by the language model.

\end{skillbox}

\subsection{Memory Skills}
\label{app:runtime_skills_memory}

The Memory category contains a single skill for converting observation and execution evidence into compact memory for the next control turn.

\begin{skillbox}{Observation Memory Summarization.}

\textbf{Purpose and Invocation.} This skill converts the current observation, segmentation and grounding results, scene memory, monitor signals, tool-execution history, and previously committed memory into compact task state for the next control turn. It uses the existing runtime memory surfaces rather than introducing an additional persistent memory representation. Its main role is to retain task-relevant facts, recent execution consequences, and unresolved uncertainty without requiring later control turns to repeatedly inspect the full raw observation and execution history.

\textbf{Inputs.} The skill operates over the existing runtime state, including committed task memory, the recent observation summary, scene memory, observation preprocessing results, prior tool calls and outcomes, robot state, and semantic tags when they are already available. Scene memory additionally provides stable object instances, task focus, temporal tracks, grounding information, and the runtime-maintained position state of each tracked instance.

\textbf{Core Procedure.}
\begin{enumerate}
    \item Collect only task-relevant facts supported by the current observation, scene memory, monitor result, tool-execution history, or previously committed memory.
    \item Preserve stable object-instance bindings across control turns when the available geometry and interaction history remain consistent, while retaining the runtime-provided position validity instead of re-deriving it from visibility alone.
    \item Summarize the consequences of recent execution, including the current target or tool binding, grounding availability, robot and gripper state, the latest tool-call outcome, repeated failures, and constraints that affect the next decision.
    \item Carry forward uncertainty explicitly when perception is weak, missing, contradictory, or retained only from history, and update committed task progress only when it is supported by observation, monitoring, or verified physical evidence.
\end{enumerate}

\textbf{Output Contract.} The skill writes into two existing text-based memory surfaces. The first, \texttt{memory\_text}, contains one concise sentence describing committed facts that remain relevant across control turns, such as completed subtasks, stable object relations, active target or tool bindings, and verified failures. The second, \texttt{recent\_observation\_summary}, provides a compact execution digest containing only evidence that may affect the next decision. The recommended structure is:

\begin{Verbatim}[breaklines=true,breakanywhere=true]
focus=target:... tool:...;
grounding=...;
robot=...;
recovery=...;
uncertainty=...
\end{Verbatim}

Metric grounding is included only when it is useful for execution or disambiguation. The skill does not create additional durable state fields such as a separate manipulation phase or execution-state schema.

\textbf{Key Constraints.}
\begin{itemize}
    \item Observation preprocessing is treated as evidence rather than as the final task-state decision.
    \item Task progress must not be inferred from an attempted action alone, and intended future actions must not be stored as completed state.
    \item A closed gripper does not establish a successful grasp without supporting physical or monitoring evidence.
    \item A visible object with missing executable grounding remains ungrounded.
    \item For tracked or temporarily missing objects, geometric usability follows the runtime-provided position state rather than visibility alone. A retained coordinate may remain usable when it is marked as valid, whereas a coordinate made uncertain by contact, failed grasp, collision, or unverified release cannot be used for grounded motion until reacquisition.
    \item Missing, weak, or contradictory evidence must remain explicit uncertainty rather than being replaced by an unsupported clean state.
    \item Stable instance identifiers or VLM-provided instance hints are used when multiple instances exist; raw segmentation rank is not used as object identity.
\end{itemize}

\textbf{Interaction with Other Components.} This skill consumes observation, grounding, monitoring, robot-state, and tool-execution evidence already maintained by the runtime and summarizes it into the memory surfaces used by subsequent control turns. In particular, it preserves the execution consequences and uncertainties that are relevant to later planning or tool-control decisions, so those decisions do not need to reconstruct the current state directly from raw observation JSON.

\end{skillbox}

\subsection{Planning Skills}
\label{app:runtime_skills_planning}

The Planning category contains five skills that cover different levels of task planning, runtime-state interpretation, monitored execution, and data-collection workflows. Table~\ref{tab:planning_skill_inventory} summarizes their responsibilities. We describe representative planning skills in greater detail in the following subsections.

\subsubsection{Planning Skill Inventory}

\begin{table}[!ht]
\centering
\begin{tabular}{p{0.24\linewidth}p{0.68\linewidth}}
\toprule
\textbf{Skill} & \textbf{Responsibility} \\
\midrule

Control Turn Planner &
Decides the next top-level action forone agent control turn from the task instruction, committed memory, observations, state information, and available runtime state. It returns the next committed memory, an executor-facing subtask, an action mode, an arm preference, and optionally a selected skill. \\

Runtime State Reasoning &
Defines how the planner interprets structured task, working, perception, scene-memory, monitor, manipulation, and recovery state. It treats structured runtime fields as the authoritative state source and specifies how that state should be used when producing the next control decision. \\

Long Horizon Execution &
Handles global tasks that must be decomposed into ordered subtasks. It refines the subtask plan, selects the next subtask, checks the previous subtask, decides whether to retry, recover, continue, or finish, and delegates concrete execution to a monitored execution boundary. \\

Monitored Subtask Execution &
Executes exactly one narrow executor-facing instruction as a monitored action unit. It monitors success, failure, stall, and timeout conditions, performs deterministic stopping or resetting when needed, and returns a structured execution result to the calling workflow. \\

EAP Data Collection &
Implements a forward/reverse EAP-style data-collection workflow. It initializes a collection run, executes forward behavior, executes reverse or reset behavior, keeps the environment reusable, and records structured run and dataset information. \\

\bottomrule
\end{tabular}
\caption{Planning skills and their responsibilities.}
\label{tab:planning_skill_inventory}
\end{table}

Table~\ref{tab:planning_skill_inventory} enumerates the complete Planning skill set. Because several skills address specialized execution or data-collection workflows, we do not describe every skill individually. Instead, the following subsection presents two representative examples that define the core planning interface.

\subsubsection{Representative Planning Skills}
\label{app:representative_planning_skills}

The following examples illustrate two complementary aspects of the planning stack: Control Turn Planner defines the decision contract for one control turn, while Runtime State Reasoning specifies how structured runtime state is interpreted when producing that decision.

\begin{skillbox}{Control Turn Planner}
\textbf{Purpose and Invocation.} The Control Turn Planner defines the planner contract for one control turn. It is used when deciding the next top-level action of a robot rollout. Given the current task state, committed memory, visual observations, state summary, and available structured runtime state, it produces the next committed memory and an executor-facing subtask.

\textbf{Inputs.} The skill receives the global task instruction, previous committed memory, the segment-start and current images, and a numeric state-summary vector. When available, it additionally receives structured runtime state containing task memory, working memory, perception results, scene memory, monitor state, recovery state, and available skills. The planner may also receive the current grasp-transport policy, per-arm manipulation state, release-guard setting, action-geometry repair policy, and runtime evaluation containing the environment's authoritative global-success signal, reward, and step counters. Under the compact planner context, scene memory is used as the single structured scene view.

\textbf{Core Procedure.}
\begin{enumerate}
    \item Determine whether any task-relevant state change has already been supported by observation, scene memory, monitor information, or environment-success evidence, and update committed memory accordingly.
    \item Read completed and failed skill histories as transitions already committed by the after-action verifier rather than independently inferring completion from planner intent.
    \item Select an actionable executor-facing subtask and one action mode from start, continue, retry, reset, replan, recover, switch, or finish. Global completion is determined by the runtime's authoritative global-success signal.
    \item Select the preferred arm from grounded geometry, end-effector poses, gripper state, and recent physical outcomes, while respecting the current manipulation state and runtime policies governing grasp transport, release, and pending action-geometry repair.
    \item For count-sensitive repeated transient actions, create one atomic event per subtask and wait for after-action verification before advancing to the next repetition.
\end{enumerate}

\textbf{Output Contract.} The skill returns JSON only using the following schema:

\begin{Verbatim}[breaklines=true,breakanywhere=true]
{
  "commit_label": "no_update | subtask_complete | state_change",
  "memory_text": "<committed task state>",
  "subtask_text": "<next executor-facing subtask>",
  "selected_skill": "<optional skill name>",
  "action_mode": "start | continue | retry | reset | replan | recover | switch | finish",
  "preferred_arm": "left | right | either",
  "semantic_tags": {}
}
\end{Verbatim}

The committed memory describes state that is already true, while the subtask field specifies the next action objective for the executor.

\textbf{Key Constraints.}
\begin{itemize}
    \item Task progress cannot be committed from planner intent alone; it must be supported by runtime evidence.
    \item A subtask-complete label only summarizes a completion already recorded by runtime and does not create a new completion transition.
    \item The global-success signal in runtime evaluation is authoritative. The planner may output finish only when this signal is true.
    \item Retry, continue, and reset require an active skill; when no active skill exists, the planner must instead start or recover with an actionable replacement subtask.
    \item Stable scene-memory instance identifiers are preferred over raw single-frame detection ranks.
    \item Arm choice is represented explicitly by the preferred-arm field and must not be encoded indirectly through task wording or object names.
    \item Manipulation state and its associated runtime policies must be preserved across replanning; confirmed or provisionally authorized attachments must not be silently rewritten into a different physical state.
    \item When action geometry is marked as relocation-pending, the selected behavior must follow the configured repair policy rather than using quarantined historical approach, grasp, or contact geometry.
\end{itemize}

\textbf{Interaction with Other Components.} The Control Turn Planner consumes the structured runtime state interpreted within the planning stack and produces the committed memory, executor-facing subtask, action mode, arm preference, and optional selected skill for the next control turn. Completion and failure transitions referenced by the planner are supplied by runtime execution history and the after-action verifier rather than created by the planner itself.

\end{skillbox}

\begin{skillbox}{Runtime State Reasoning}
    
\textbf{Purpose and Invocation.} Runtime State Reasoning defines how structured \texttt{agent\_state} should be interpreted inside a control turn. Its role is to make the structured runtime fields the authoritative source for task, working, perception, scene, monitor, manipulation, and recovery state, while using raw images and summaries as supporting evidence rather than reconstructing long execution histories from them.

\textbf{Inputs.} The reasoning payload exposes three main groups of state. Task memory contains the global task, committed memory, committed facts, execution plan, completed and failed skills, and the runtime task-finished flag. Working memory contains the active skill and instruction, recent observation information, scene memory, recent tool calls, local progress information, recovery history, manipulation state, runtime policy flags, and semantic tags. Monitor and recovery state provide the current rollout status, progress and failure information, together with retry, reset, and replan budgets or pending recovery actions.

When compact planner context is enabled, scene memory is the single planner-visible scene source. Public grasp/contact availability and manipulation state remain visible there, while complete masks, candidate poses, and attachment transforms remain private to runtime.

\textbf{Core Procedure.}
\begin{enumerate}
    \item Inspect structured runtime fields first and use raw images or observation summaries as supporting evidence rather than reconstructing long histories independently.
    \item Use scene memory for stable object identity and spatial grounding, and interpret each instance's position state independently of visibility. Current-verified geometry supports the normal grounded contract, memory-valid geometry permits only a safe approach before fresh verification, and motion-uncertain geometry forbids use of historical coordinates.
    \item Use monitor and recovery state when deciding whether execution should continue, retry, recover, replan, or finish.
    \item Interpret manipulation state together with the configured grasp-transport and release policies. Confirmed holding, provisional holding, pending release verification, and release-recovery-required are kept as distinct runtime states with different allowed next actions.
    \item Use public operation targets for placement decisions. Reference-region targets are selected from runtime-computed relations rather than recomputing their metric coordinates in the planner.
    \item Keep committed memory concise and restricted to facts that are already true; when perception conflicts with previous memory, preserve the uncertainty instead of silently overwriting it.
\end{enumerate}

\textbf{Output Contract.} This skill defines the structured payload that the planner should inspect rather than a separate planner-output schema. The control-turn payload follows the contract:

\begin{Verbatim}[breaklines=true,breakanywhere=true]
{
  "global_task": "...",
  "trigger": "...",
  "agent_state": {},
  "available_skills": [],
  "memory_harness": {},
  "runtime_state_reasoning": {}
}
\end{Verbatim}

The planner is instructed to inspect \texttt{agent\_state} first and then use the memory harness and available skills when producing the next control decision.

\textbf{Key Constraints.}
\begin{itemize}
    \item Structured runtime fields are the authoritative state source.
    \item Scene memory is used for stable instance identity and spatial grounding; private candidate identifiers are not exposed as planner choices.
    \item Verified physical roles must not be silently relabeled by a later single semantic observation.
    \item Historical coordinates marked motion-uncertain cannot be used for grounded motion.
    \item Under safe-motion repair, only runtime-validated open-gripper retreat, runtime-derived safe-height motion, and fresh camera observation are permitted while action geometry is pending.
    \item Transport authorization does not by itself imply confirmed attachment; provisional and confirmed holding states remain distinct.
    \item A released object awaiting verification must not immediately enter the next manipulation, while a verified released-target miss requires reacquisition rather than continued observation-only verification.
    \item Committed memory must remain short and contain only state that is already true; working-memory details should not be copied wholesale into persistent memory.
\end{itemize}

\textbf{Interaction with Other Components.} Runtime State Reasoning defines how the structured state supplied to the planning stack should be read before the next control decision is produced. The Control Turn Planner then uses this interpreted task, scene, monitor, manipulation, and recovery state together with the available skills and memory harness to determine the next control-turn output.

\end{skillbox}

\subsection{Monitoring Skills}
\label{app:runtime_skills_monitoring}

The Monitoring category contains a single skill for semantic OOD and failure-state detection during VLA rollout execution.

\begin{skillbox}{OOD Detection}
    
\textbf{Purpose and Invocation.} This skill judges whether the currently executing subtask has entered a semantic out-of-domain, blocked, failed, or visually invalid state. It belongs to the monitoring path rather than the execution path: the skill classifies execution health and returns an explicit structured OOD result, but does not execute recovery actions, replace progress or success detection, or directly determine low-level robot control. The classification is made relative to the active subtask rather than only to the global task.

\textbf{Inputs.} The skill receives the active subtask, the latest observation summary, current monitor status, recovery state, structured execution context, and the current and maximum environment step counts. The execution context may provide progress, task-success, and action-chunk information. The complete payload is evaluated jointly by the VLM; the runtime does not infer specific OOD classes afterward from a free-text summary.

\textbf{Core Procedure.}
\begin{enumerate}
    \item Read the active subtask and establish what the VLA is currently expected to accomplish.
    \item Inspect the latest observation, monitor status, recovery state, step budget, and execution context together. Explicit evidence such as target disappearance, grasp loss, motion blockage, camera failure, or a contradictory scene state is preferred over generic wording.
    \item Select exactly one primary OOD scenario from the supported vocabulary, choosing the narrowest class justified by the available evidence.
    \item Return the classification directly together with a short evidence-based reason and, when useful, consistent structured monitor signals.
\end{enumerate}

\textbf{Output Contract.} The skill returns one JSON object. The primary contract is the explicit \texttt{OOD\_scenario} field; supporting monitor signals are optional.

\begin{Verbatim}[breaklines=true,breakanywhere=true]
{
  "status": "ok",
  "selected_skill": "ood-detection",
  "OOD_scenario":
    "none | object_not_visible | motion_blocked |
     grasp_lost | scene_drift_detected | requires_replan",
  "reason": "<short evidence-based explanation>",
  "confidence": "<0.0--1.0, optional>",
  "signals": [
    {
      "name": "<runtime-aligned signal>",
      "level": "info | warning | error",
      "reason": "<short explanation>",
      "score": "<0.0--1.0>",
      "details": {}
    }
  ],
  "analysis_note": "<optional concise explanation>"
}
\end{Verbatim}

The supported primary scenarios are \texttt{none}, \texttt{object\_not\_visible}, \texttt{motion\_blocked}, \texttt{grasp\_lost}, \texttt{scene\_drift\_detected}, and \texttt{requires\_replan}. If supporting signals are returned, they must remain consistent with the primary classification.

\textbf{Key Constraints.}
\begin{itemize}
    \item Classification is conservative: when the available payload does not support a confident OOD diagnosis, the skill returns \texttt{none}.
    \item The skill must not fabricate scene facts, object identities, grasp states, or failure causes.
    \item It must output the OOD scenario explicitly rather than returning only a natural-language summary and relying on runtime post-processing to classify it.
    \item Conditions already handled by progress monitoring or task-success logic should not be forced into an OOD class.
    \item The narrowest justified failure class is preferred over a generic escalation label.
    \item Recovery tool plans and tool-call arguments are outside the scope of this skill.
    \item When evidence conflicts, the classification follows the strongest explicit evidence; \texttt{requires\_replan} is used only when the payload supports such escalation.
\end{itemize}

\textbf{Interaction with Other Components.} The monitoring runtime constructs the evaluation payload and invokes this skill during VLA rollout. Its structured OOD result is consumed by the monitoring evaluator, and the returned signals are then available to the agent runtime when deciding whether VLA control should continue or be handed off to the tool-calling path. The skill itself performs only semantic execution-health classification and does not generate the subsequent tool calls.

\end{skillbox}

\subsection{Tool Calling Skills}
\label{app:runtime_skills_tool_calling}

The Tool Calling category contains 19 skills that structure the use of runtime manipulation tools. They are organized into one routing skill, ten reusable primitive skills, and eight workflow skills. The routing skill selects an appropriate workflow and post-execution intent; primitive skills specify the invocation contract of individual runtime tools; and workflow skills construct short, situation-dependent sequences from the available primitives.

\subsubsection{Tool-Calling Skill Inventory}

\begin{table}[t]
\centering
\begin{tabular}{p{0.27\linewidth}p{0.65\linewidth}}
\toprule
\textbf{Skill} & \textbf{Responsibility} \\
\midrule

Tool-Calling Router &
Selects one workflow and a post-execution intent from an explicit OOD scenario or monitor signal. It returns a structured routing decision and does not itself execute tools. \\

Close Gripper &
Closes a selected gripper to re-establish or stabilize grasp state after an uncertain, failed, or slipping grasp. \\

Contact Displace &
Applies a very small bounded end-effector displacement when local contact should be released or probed without initiating a new task-level action. \\

Lift End Effector &
Creates bounded vertical end-effector clearance when the arm or gripper is locally blocked or obstructed. \\

Move EE to Grounded Instance &
Moves an end effector toward manipulation geometry associated with a grounded scene-memory instance or a public placement target. Executable pose selection remains runtime-side. \\

Move EE to Pose &
Moves an end effector toward an already validated absolute world-frame target pose or position. It does not perform semantic grounding. \\

Move to Home &
Moves one or both end effectors toward the configured original or home pose before retry, replanning, or termination of an unsafe local rollout. \\

Open Gripper &
Opens a selected gripper to release an unstable or failed grasp state or to prepare for safe retreat. \\

Reobserve Scene &
Refreshes visual and contextual observation after an abnormal rollout state or after another tool changes robot posture or scene visibility. \\

Retreat Arm &
Creates local end-effector clearance after blocked motion, failed grasp, uncertain contact, or a stalled posture. \\

Safe Reset Posture &
Moves the end effector by a bounded step toward a conservative original or home posture when local recovery motion is insufficient. \\

\bottomrule
\end{tabular}
\caption{Routing and primitive skills in the Tool Calling category.}
\label{tab:tool_calling_router_primitives}
\end{table}

\begin{table}[!ht]
\centering
\begin{tabular}{p{0.27\linewidth}p{0.65\linewidth}}
\toprule
\textbf{Skill} & \textbf{Responsibility} \\
\midrule

Go Home and Retry &
Handles exhausted or unproductive rollouts by constructing a situation-specific tool sequence that returns the robot toward a reusable posture before control is returned to planning. \\

Recover Grasp Lost &
Constructs a bounded tool sequence when the manipulated object is no longer attached to or controlled by the gripper. \\

Recover Motion Blocked &
Constructs a situation-specific tool sequence when local contact or blockage prevents the current motion from continuing safely. \\

Recover Object Not Visible &
Constructs a tool-use plan that improves observability or returns control to planning when a task-relevant object cannot be reliably observed. \\

Recover Requires Replan &
Returns control to planning when the current subtask is no longer valid. It may produce no physical tool calls when physical intervention is unnecessary. \\

Recover Scene Drift &
Refreshes observation and determines whether execution can be retried or should be replanned when the scene has changed enough to invalidate the current rollout context. \\

Retreat and Reobserve &
Handles a locally stalled but potentially retryable subtask by constructing a short tool sequence for clearance and refreshed observation. \\

Task-Level Tool Control &
Uses grounded runtime tools as the temporary low-level controller to make observable progress on the global instruction under the dedicated task-level control mode. \\

\bottomrule
\end{tabular}
\caption{Workflow skills in the Tool Calling category. Each workflow constructs a situation-dependent sequence from the runtime tools currently available.}
\label{tab:tool_calling_workflows}
\end{table}

Table~\ref{tab:tool_calling_router_primitives} and Table~\ref{tab:tool_calling_workflows} enumerate the complete Tool Calling skill set. To avoid repeating similar invocation and output contracts across closely related skills, we do not describe every skill individually. Instead, the following subsection presents three representative examples spanning the routing, primitive, and workflow levels.

\subsubsection{Representative Tool-Calling Skills}
\label{app:representative_tool_calling_skills}

\begin{skillbox}{Tool-Calling Router}

\textbf{Purpose and Invocation.}
The Tool-Calling Router is invoked when the runtime needs to select an appropriate tool-use workflow from the current execution state. When invoked from the monitoring path, it may consume an explicit OOD scenario or monitor signal. Its role is to select one available workflow and determine the post-execution intent. The router itself does not execute tools or generate low-level tool calls.

\textbf{Inputs.} The router receives the monitor signal or OOD scenario, its supporting reason, the current subtask, the current observation summary, recovery state, previous tool-calling history, and the set of workflows currently available to the runtime.

\textbf{Core Procedure.}
\begin{enumerate}
    \item Read the monitor or OOD signal without re-classifying the raw observation.
    \item Select exactly one available workflow that best matches the current signal, subtask, and prior tool-calling history. The least disruptive workflow that can safely restore control is preferred.
    \item Avoid repeating a workflow that has already failed unless new evidence justifies another attempt.
    \item Select a post-execution intent: retry the current subtask, return control to planning, or abort when the state is unsafe or repeated recovery has failed.
\end{enumerate}

\textbf{Output Contract.} The router returns JSON only. The current routing contract is:

\begin{Verbatim}[breaklines=true,breakanywhere=true]
{
  "selected_skill": "tool-calling-router",
  "OOD_scenario": "grasp_lost",
  "tool_workflow": "recover-grasp-lost",
  "post_execution_intent": "retry",
  "reason": "<short routing rationale>"
}
\end{Verbatim}

\textbf{Key Constraints.}
\begin{itemize}
    \item The router assumes that monitoring has already produced the failure signal and does not re-classify raw images.
    \item It selects exactly one workflow and does not output low-level tool calls.
    \item The selected workflow must appear in the runtime-provided workflow list.
    \item If the signal is ambiguous, the router prefers a workflow that refreshes observation or returns control to planning.
    \item If no workflow fits, the fallback is to return control to planning through the replanning workflow.
    \item Repeated failed physical recovery should lead to replanning or abort rather than unconditional repetition.
\end{itemize}

\textbf{Interaction with Other Components.} The router consumes the explicit execution-health signal produced by monitoring and selects the workflow that should handle the current situation. The selected workflow subsequently generates concrete tool calls. Runtime validates the selected workflow before invoking the workflow planner.

\end{skillbox}

\begin{skillbox}{Move End Effector to Grounded Instance}

\textbf{Purpose and Invocation.} This primitive is used when a tool-calling procedure needs to move an end effector toward manipulation geometry that is already represented in scene memory. For grasp and contact operations, the planner selects a grounded scene instance. For placement, it selects a public operation target, while runtime binds the transport-authorized held object and resolves the executable arm-specific pose. The planner therefore selects semantic grounding, whereas private pose-candidate selection remains inside the runtime.

\textbf{Inputs.} The primitive uses scene memory, public placement targets, the latest perception evidence, robot state, the current subtask, previous tool-call outcomes, the runtime tool allowlist, and the configured grasp-transport policy. Scene instances may contain grounded operation-pose candidates, but private candidate identifiers are not exposed as planner tool arguments. Public placement targets instead expose stable target identifiers, target type, support and occupancy state, target coordinates, and executable arm choices.

\textbf{Core Procedure.}
\begin{enumerate}
    \item Determine the manipulation mode. Portable objects intended for later carry or placement use grasp mode, whereas press, push, brace, or articulated interactions use contact mode. Placement uses the dedicated place mode.
    \item Select a grounded public reference. Grasp and contact may use an exact scene-instance identity; place uses an exact public operation-target identifier. When identity binding is required, role-based or focus-based selection cannot replace the exact public identity.
    \item Select the appropriate grounded point and motion contract. A new portable-object grasp uses the grounded approach pose as the mandatory staging pose before the final grasp pose and gripper closure. Contact uses separately grounded contact geometry. Placement uses the runtime-provided placement target and may first move to its approach pose before issuing the final place motion.
    \item Pass the public grounding request to runtime. Runtime resolves the matching private candidate, validates the grounded position and orientation, applies bounded motion limits, and executes the resulting end-effector command.
    \item Fail closed when the required identity, finite 3D grounding, position validity, placement support, occupancy state, or transport authority is unavailable.
\end{enumerate}

\textbf{Output Contract.} This skill produces a structured invocation of the runtime tool \texttt{move\_ee\_to\_grounded\_instance}. The principal arguments are shown below; the runtime interface additionally supports bounded offsets, clearance parameters, focus-based selection when permitted, and gripper preconditions.

\begin{Verbatim}[breaklines=true,breakanywhere=true]
{
  "tool_name": "move_ee_to_grounded_instance",
  "args": {
    "arm": "left | right",
    "instance_id": "<public scene instance, optional>",
    "action_mode": "grasp | contact | place",
    "target_id": "<public operation target; required for place>",
    "point_key": "<grounded point type>",
    "held_instance_id": "<held instance, optional>",
    "preserve_height": "<boolean, optional>",
    "gripper_precondition": "<requested state, optional>",
    "target_quat_wxyz": "grounded | preserve | <quaternion>",
    "max_translation": "<bounded translation>",
    "steps": "<bounded step count>"
  },
  "reason": "<short tool-call rationale>"
}
\end{Verbatim}

The runtime result follows the common \texttt{RecoveryToolResult} interface. For audit, the runtime may report the resolved operation candidate, operation target, action mode, and geometry source, while the planner never selects private candidate identifiers.

\textbf{Key Constraints.}
\begin{itemize}
    \item The primitive must use grounded scene-memory geometry and must not invent coordinates.
    \item Private operation-candidate identifiers never enter planner tool arguments.
    \item When exact identity binding is required, the public scene-instance identifier must be supplied explicitly.
    \item A scene instance with motion-uncertain position state cannot be used through its retained historical coordinate.
    \item A new portable-object grasp must first reach its grounded approach pose before proceeding to its grasp pose and gripper closure, unless recovery history already establishes that the same arm reached that approach for the same grasp attempt.
    \item Contact interactions with anchored or articulated mechanisms remain contact operations even when the gripper may subsequently close around a handle.
    \item Place mode uses a public operation-target identifier rather than a destination instance identifier. Runtime recomputes support and occupancy before execution and rejects stale or occupied targets.
    \item Missing or ambiguous visual evidence does not cause this primitive to insert diagnostic motion, re-observation, gripper opening, retreat, or return-to-grasp actions automatically; such actions must be requested separately when needed.
\end{itemize}

\textbf{Interaction with Other Components.} This primitive is called by a tool-calling workflow when grounded end-effector motion is required. It consumes public grounding already stored in scene memory and leaves private pose resolution, bounded execution, and validation to the runtime. If an instance cannot be resolved or its geometry is no longer valid, the skill directs the calling procedure toward re-observation or replanning rather than falling back to guessed coordinates.

\end{skillbox}

\begin{skillbox}{Recover Grasp Lost}

\textbf{Purpose and Invocation.} This workflow is used when monitoring has already identified a lost or unstable grasp, meaning that the manipulated object appears no longer attached to or controlled by the gripper. It is a planning harness for tool use rather than a fixed recovery script: the workflow constructs a situation-specific sequence from the tools currently available and remains within the supplied tool budget.

\textbf{Inputs.} The workflow receives the grasp-loss monitor signal, the interrupted subtask, an observation summary containing object visibility, gripper state, and robot posture, previous recovery tool calls and outcomes, the runtime-provided tool set, the maximum tool-call budget, and the allowed post-recovery intents.

\textbf{Core Procedure.}
\begin{enumerate}
    \item Assess whether the object remains visible and reachable, whether the gripper is still closed on anything, and whether the arm is near a collision or obstruction.
    \item Select only the tools needed for the observed state. The workflow may open the gripper to clear an unstable grasp, retreat or lift the arm to create clearance, move toward an already grounded instance for bounded retry setup, or refresh observation after physical recovery.
    \item Repeat observation only when an intermediate physical tool has changed the scene or robot posture.
    \item Select retry when the target remains visible and reachable after recovery, replan when the target moved or the original subtask is no longer valid, and abort when recovery tools fail or the scene is unsafe.
\end{enumerate}

\textbf{Output Contract.} The workflow returns JSON only. It produces an ordered list of situation-specific tool calls, together with a post-recovery intent and an explicit stop condition.

\begin{Verbatim}[breaklines=true,breakanywhere=true]
{
  "selected_workflow": "recover-grasp-lost",
  "tool_calls": [
    {
      "tool_name": "<available runtime tool>",
      "args": {},
      "reason": "<reason for this tool call>"
    }
  ],
  "post_recovery_intent": "retry | replan | abort",
  "stop_condition": "<workflow stop condition>"
}
\end{Verbatim}

The repository provides reference plans such as opening the gripper, retreating, and re-observing, but these plans are examples rather than mandatory sequences.

\textbf{Key Constraints.}
\begin{itemize}
    \item The workflow assumes that monitoring has already classified grasp loss and does not perform that classification itself.
    \item It does not decide whether the global task or subtask has succeeded; it only prepares the runtime for retry, replanning, or abort.
    \item Only tools present in the runtime-provided tool list may be used, and the complete plan must respect the supplied tool budget.
    \item Uncertain object or arm state should not trigger repeated grasp-related physical recovery without new evidence.
    \item When useful for safety, an unstable grasp should be released before a large retreat motion.
    \item Grounded-instance motion is used only when scene memory already contains the finite 3D instance required for the retry setup.
    \item If opening the gripper fails, repeated forceful motion is avoided; if retreat fails, lifting is attempted only when that tool is available and safe.
    \item If the target is no longer visible after re-observation, the workflow returns control to replanning.
\end{itemize}

\textbf{Interaction with Other Components.} Monitoring identifies the grasp-loss condition, after which the routing layer may select this workflow. Recover Grasp Lost then composes concrete calls from the runtime-provided primitive tools, such as gripper control, retreat, lifting, grounded motion, and re-observation. Its output returns either retry, replan, or abort as the intended control state after the tool sequence.

\end{skillbox}

\subsection{Effect Verification Skills}
\label{app:runtime_skills_effect_verification}

The Effect Verification category contains a single runtime skill for assessing the immediate physical outcome of an executed action or tool sequence from before-and-after evidence. This runtime verification is distinct from the AFK evidence verdict in Section~3.4: the former evaluates the current execution and subtask transition, whereas the latter binds an execution event to persistent knowledge and classifies its evidence as support, opposition, or unverified.

\begin{skillbox}{Action Effect Verification}

\textbf{Purpose and Invocation.} This skill is invoked after a tool sequence has executed and a fresh observation is available. It determines whether the observed physical effect is consistent with the current subtask and the intended action. The verifier is the after-action authority for the current subtask transition: a pre-execution post-action intent is treated only as a hint, while the resulting subtask status is determined from fresh evidence. The verifier does not introduce a new runtime memory schema; its result is compressed back into the existing recovery history and recent observation summary.

\textbf{Inputs.} The verifier receives the global task, current subtask, proposed post-execution intent, expected outcome, reason for the preceding tool sequence, and the executed tool calls together with their arguments and runtime results. It additionally receives before- and after-action evidence containing robot state, scene memory, observation summaries, and runtime evaluation, as well as recent recovery history. For placement operations, the runtime may additionally provide deterministic placement validation covering support and occupancy, target position, gripper state, end-effector clearance, and cross-observation stability.

\textbf{Core Procedure.}
\begin{enumerate}
    \item Infer the intended physical effect from the current subtask and executed tool sequence rather than from task-specific object-class rules or planner prose.
    \item Compare before- and after-action evidence using current grounded geometry, track continuity, robot and gripper state, monitor signals, and measured state changes. Current object positions are taken from the current observation fields; history-smoothed positions are used for identity continuity rather than measuring the displacement caused by the latest action.
    \item Apply effect-specific verification when required. For grasp or carry, compare object motion with end-effector motion and their relative coupling. For placement, require the runtime placement validator to report a verified placement before accepting release or subtask completion. For contact-based transient actions, verify a complete grounded contact-and-release cycle from the same arm rather than inferring repetitions from controller steps or traveled distance.
    \item Classify the action effect as verified, contradicted, or unverified according to the available evidence.
    \item Determine the current subtask status from fresh physical evidence and select the corresponding control recommendation for continued execution, retry, or replanning.
\end{enumerate}

\textbf{Output Contract.} The verifier returns JSON only and never returns additional tool calls.

\begin{Verbatim}[breaklines=true,breakanywhere=true]
{
  "effect_verified": "true | false | unverified",
  "effect_type":
    "grasp | place | release | push | press | move |
     align | open | close | cover | uncover | unknown",
  "confidence": 0.0,
  "evidence_summary": "<short evidence-based summary>",
  "failure_reason": "",
  "next_constraint": "<constraint for the next plan>",
  "memory_update": "<compact recovery-history update>",
  "subtask_status":
    "in_progress | completed | failed | uncertain",
  "recommended_control": "continue | retry | replan"
}
\end{Verbatim}

The memory update is kept short and action-oriented rather than copying raw observations or scene-memory records into persistent memory.

\textbf{Key Constraints.}
\begin{itemize}
    \item Successful tool execution is not sufficient evidence that the intended physical effect occurred.
    \item A closed gripper or a successful close command does not by itself establish that an object is being held.
    \item Perception-query metadata, task wording, and planner rationales describe intent and must not be used as direct evidence for physical relations such as holding, contact, release, coverage, or clearance.
    \item Current action effects are measured from current before-and-after grounding. Historical reference poses and history-smoothed positions must not substitute for the current object position when measuring instantaneous motion.
    \item Placement is verified only when the runtime placement validator reports \texttt{verified=true}. A successful final move or gripper-open command alone is insufficient.
    \item If placement validation reports that recovery is required after release, the subtask is treated as failed and control is returned for replanning rather than repeatedly requesting observation.
    \item An empty tool sequence with a direct replanning intent does not imply that a new physical effect occurred. The verifier evaluates whether current evidence supports the claimed handoff state.
    \item One verified transient contact cycle counts as exactly one event; controller steps, internal repetitions, or traveled distance do not create additional verified repetitions.
    \item \texttt{subtask\_status=completed} requires fresh evidence of the current subtask's physical stop condition. Requested intent or successful tool return is insufficient.
    \item Missing, occluded, contradictory, or insufficient evidence produces an \emph{unverified} result rather than being forced into success or failure.
    \item The runtime environment-success signal remains authoritative for global task completion and may bypass this verifier once global success has already been established.
\end{itemize}

\textbf{Interaction with Other Components.}
This skill is applied after an action or tool sequence has executed and fresh post-action evidence has been collected. Its immediate result is used by the runtime to update the local subtask state and determine whether execution should continue, retry, or return to planning. The verification record may also become part of the execution event subsequently used by the AFK evidence-assessment pipeline; AFK maintenance separately determines whether that event supports, opposes, or leaves unverified a persistent knowledge entry.

\end{skillbox}

\end{document}